%% file: MMSE_NMF.tex
\pdfoutput=1
\documentclass[preprint,authoryear,review,12pt]{elsarticle}
\usepackage{amsmath}
\usepackage{graphics}
\usepackage{multirow}
\include{def_bold}

\usepackage{multicol}
\usepackage{color}
\usepackage{graphics}
\newfont{\rsfsten}{rsfs10 scaled 1200}
\newfont{\rsfsseven}{rsfs10 scaled 1200}
\newfont{\rsfsfive}{rsfs10 scaled 1200}

\begin{document}

\begin{frontmatter}




\title{Source Separation using Regularized NMF with MMSE Estimates under GMM Priors with Online Learning for The Uncertainties}


\author{Emad M. Grais and Hakan Erdogan\\
\{grais, haerdogan\}@sabanciuniv.edu}

\address{Faculty of Engineering and Natural Sciences,\\
Sabanci University, Orhanli Tuzla, 34956, Istanbul.}

\begin{abstract}
We propose a new method to enforce priors on the solution of the nonnegative matrix factorization (NMF). The proposed algorithm can be used for denoising or single-channel source separation (SCSS) applications. The NMF solution is guided to follow the Minimum Mean Square Error (MMSE) estimates under Gaussian mixture prior models (GMM) for the source signal. In SCSS applications, the spectra of the observed mixed signal are decomposed as a weighted linear combination of trained basis vectors for each source using NMF. In this work, the NMF decomposition weight matrices are treated as a distorted image by a distortion operator, which is learned directly from the observed signals. The MMSE estimate of the weights matrix under GMM prior and log-normal distribution for the distortion is then found to improve the NMF decomposition results. The MMSE estimate is embedded within the optimization objective  to form a novel regularized NMF cost function. The corresponding update rules for the new objectives are derived in this paper. Experimental results show that, the proposed regularized NMF algorithm improves the source separation performance compared with using NMF without prior or with other prior models.
\end{abstract}


\end{frontmatter}


 
%
\section{Introduction}
\label{sec:intro}
Nonnegative matrix factorization~\citep{Lee:01:afnmf} is an important tool for source separation applications, especially when only one observation of the mixed signal is available. NMF is used to decompose a nonnegative matrix into a multiplication of two nonnegative matrices, a basis matrix and a gains matrix. The basis matrix contains a set of basis vectors and the gains matrix contains the weights corresponding to the basis vectors in the basis matrix. The NMF solutions are found by solving an optimization problem based on minimizing a predefined cost function. As most optimization problems, the main goal in NMF is to find the solutions that minimize the cost function without considering any prior information rather than the nonnegativity constraint. There have been many works that tried to enforce prior information related to the nature of the application on the NMF decomposition results. For audio source separation applications, the continuity and sparsity priors were enforced in the NMF decomposition weights~\citep{Virtanen:07:msssbnmfwtcasc}. In~\citet{nancy:09:ehsbnmf}, and \citet{nancy:10:ehsbnmf}, smoothness and harmonicity priors were enforced on the NMF solution in Bayesian framework and applied to music transcription. In \citet{Wilson:08:sdunmfwp}, and \citet{kevin:08:srnmfwtdfs} the regularized NMF was used to increase the NMF decomposition weights matrix likelihood under a prior Gaussian distribution. In \citet{Fevotte:09:nmfwisd}, Markov chain prior model for smoothness was  
used within a Bayesian framework in regularized NMF with Itakura-Saito (IS-NMF) divergence. In \citet{virtanen:08:betnnffasm}, the conjugate prior distributions on the NMF weights and basis matrices solutions with the Poisson observation model within Bayesian framework was introduced. The Gamma distribution and the Gamma Markov chain \citep{taylan:07:cgmrfmns} were used as priors for the basis and weights/gains matrices respectively in \citet{virtanen:08:betnnffasm}. A mixture of Gamma prior model was used as a prior for the basis matrix in \citet{virtanen:09:mogpfnnmfbss}. The regularized NMF with smoothness and spatial decorrelation constraints was used in \citet{Chen:06:cnmfmfeaiedoad} for EEG applications. In \citet{cichocki:06:nafnmfiatbss}, and \citet{Chen:06:cnmfmfeaiedoad}, a variety of constrained NMF algorithms were used for different applications. 

In supervised single channel source separation (SCSS), NMF is used in two main stages, the training stage and the separation stage \citep{Schmidt:06:scssusnnmf,emad:11:aossbinmffscsms,emad:11:scsmsunmfasm,emad:11:scsmsunmfwswasm,emad:12:avsrwbmscss,emad:12:stpsnmfscss}. In the training stage, NMF is used to decompose the spectrogram of clean training data for the source signals into a multiplication of trained basis and weights/gains matrices for each source. The trained basis matrix is used as a representative model for the training data of each source and the trained gains matrices are usually ignored. In the separation stage, NMF is used to decompose the mixed signal spectrogram as a nonnegative weighted linear combination of the columns in the trained basis matrices. The spectrogram estimate for each source in the mixed signal can be found by summing its corresponding trained basis terms from the NMF decomposition during the separation stage. One of the main problems of this framework is that, the estimate for each source spectrogram is affected by the other sources in the mixed signal. The NMF decomposition of the weight combinations in the separation stage needs to be improved. To improve the NMF decomposition during the separation stage, prior information about the weight combinations for each source can be considered.

In this work, we introduce a new method of enforcing the NMF solution of the weights matrix in the separation stage to follow certain estimated patterns. We assume we have prior statistical informations about the solution of the NMF weights matrix. The Gaussian mixture model (GMM) is used as a prior model for the valid/expected weight combination patterns that can exist in the columns of the weights matrix that are related to the nature of the source signals. 
Here, in the training stage, NMF is also used to decompose the spectrogram of the training data into trained basis and weights/gains matrices for each source. In this work, the trained gains matrix is used along with the trained basis matrix to represent each source. We can see the columns of the trained gains matrix as valid weight combinations that their corresponding bases in the basis matrix can jointly receive for a certain type of source signal. The columns of the trained gains matrix can be used to train a prior model that captures the statistics of the valid weight combinations that the bases can receive. The prior gain model and the trained basis matrix for each source can be used to represent each source in the separation stage. During the separation stage, the prior model can guide the NMF solution to prefer these valid weight patterns. The multivariate Gaussian mixture model (GMM) can be used to model the trained gains matrix \citep{grais:12:rnmfgmpsscss}. The GMM is a rich model which captures the statistics and the correlations of the valid gain combinations for each source signal. GMMs are extensively used in speech processing applications like speech recognition and speaker verification. GMMs are used to model the multi-modal nature in speech feature vectors due to phonetic differences, speaking styles, gender, accents \citep{rabiner:93:fos}. We are conjecturing that the weight vectors of the NMF gains matrix can be considered as a feature extracted from the signal in a frame so that it can be modeled well with a GMM. The columns in the trained weights matrix are normalized, and their logarithm is then calculated and used to train the GMM prior. The basis matrix and the trained GMM prior model for the weights are jointly used as a trained representative model for the source training signals.  

In the separation stage and after observing the mixed signal, NMF is used again to decompose the mixed signal spectrogram as a weighted linear combination of the trained basis vectors for the sources that are involved in the observed mixed signal. The conventional NMF solution for the weight combinations is found to minimize a predefined NMF cost function ignoring that, for each set of trained basis vectors of a certain source signal there is a set of corresponding valid weight combinations that the bases can possibly receive. In \citet{grais:12:rnmfgmpsscss}, the prior GMM that models the valid weight combinations for each source is used to guide the NMF solution for the gains matrix during the separation stage. The priors in \citet{grais:12:rnmfgmpsscss} are enforced by maximizing the log-likelihood of the NMF solution with the trained prior GMMs. The priors in \citet{grais:12:rnmfgmpsscss} are enforced without evaluating how good the NMF solution is without using the priors. For example, if the NMF solution without prior is not satisfactory, we would like to rely more on the priors and vice versa. 

In this work, we introduce a new strategy of applying the priors on the NMF solutions of the gain matrix during the separation stage. The new strategy is based on evaluating how much the solution of the NMF gains matrix needs to rely on the prior GMMs. The NMF solutions without using priors for the weights matrix for each source during the separation stage can be seen as a deformed image, and its corresponding valid weights/gains matrix is needed to be estimated under the GMM prior. The deformation operator parameters which measure the uncertainty of the NMF solution of the weights matrix are learned directly from the observed mixed signal. The uncertainty in this work is a measurement of how far the NMF solution of the weights matrix during the separation stage is from being a valid weight pattern that is modeled in the prior GMM. The learned uncertainties are used with the minimum mean square error (MMSE) estimator to find the estimate of the valid weights matrix. The estimated valid weights matrix should also consider the minimization of the NMF cost function. To achieve these two goals, a regularized NMF is used to consider the valid weight patterns that can appear in the columns of the weights matrix while decreasing the NMF cost function. The uncertainties within MMSE estimates of the valid weight combinations are embedded in the regularized NMF cost function for this purpose. The uncertainty measurements play very important role in this work as we will show in next sections. If the uncertainty of the NMF solution of the weights matrix is high, that means the regularized NMF needs more support from the prior term. In case of low uncertainty, the regularized NMF needs less support from the prior term. Including the uncertainty measurements in the regularization term using MMSE estimate makes the proposed regularized NMF algorithm decide automatically how much the solution should rely on the prior GMM term. This is the main advantage of the proposed regularized NMF compared to the regularization using the log-likelihood of the GMM prior or other prior distributions \citep{emad:12:gmgprnmfscss,grais:12:rnmfgmpsscss,canny:04:gap}. Incorporation of the uncertainties that measure the extent of distortion in the NMF weights matrix solutions in the regularization term is a main novelty of this work, which has not been seen before in the regularization literature.     

The remainder of this paper is organized as follows: In section~\ref{NMF}, we give a brief explanation about NMF. In section~\ref{sp}, we discuss the problem of single channel source separation and its formulation. In Section \ref{conv_nmf}, we show the conventional usage of NMF in SCSS problems. Section \ref{motivation_RNMF} describes the needs for a regularized NMF. Sections \ref{motivation_RNMF2} to \ref{r_CNMF} introduce the new regularized NMF and how it is used in the SCSS problem, which is the main contribution of this paper. Section \ref{masks} indicates the source signal reconstruction after NMF decomposition. In the remaining sections, we present our observations and the results of our experiments.

\section{Nonnegative matrix factorization}
\label{NMF}
Nonnegative matrix factorization is used to decompose any nonnegative matrix $\bV$ into a multiplication of a nonnegative basis matrix $\bB$ and a nonnegative gains or weights matrix $\bG$ as follows:
\begin{equation}
\label{general_NMF}
	\bV\approx\bB\bG.
\end{equation}
The columns of matrix $\bB$ contain nonnegative basis or dictionary vectors that are optimized to allow the data in $\bV$ to be approximated as a nonnegative linear combination of its constituent vectors. Each column in the gains/weights matrix $\bG$ contains the set of weight combinations that the basis vectors in the basis matrix have for its corresponding column in the $\bV$ matrix. To solve for matrix $\bB$ and $\bG$, different NMF cost functions can be used. For audio source separation applications, the Itakura-Saito (IS-NMF) divergence cost function~\citep{Fevotte:09:nmfwisd} is usually used. This cost function is found to be a good measurement for the perceptual differences between different audio signals \citep{Fevotte:09:nmfwisd,xabier:11:assdnmfss}. The IS-NMF cost function is defined as:
\begin{equation}
\label{div_is}
\min_{\bB,\bG} D_{IS}\left(\bV\left|\right|\bB\bG\right),
\end{equation}
where 
\[
D_{IS}\left(\bV\left|\right|\bB\bG\right)=\sum_{m,n}\left(\frac{\bV_{m,n}}{\left(\bB\bG\right)_{m,n}}-\log\frac{\bV_{m,n}}{\left(\bB\bG\right)_{m,n}}-1\right).
\] 
The IS-NMF solutions for equation~(\ref{div_is}) can be computed by alternating multiplicative updates of $\bB$ and $\bG$ \citep{Fevotte:09:nmfwisd,xabier:11:assdnmfss} as:
\begin{equation}
\label{basis_is}
\bB\leftarrow \bB\otimes\frac{\frac{\bV}{\left(\bB\bG\right)^2}\bG^T}{\frac{\Bone}{\bB\bG}\bG^T},	
\end{equation} 
\begin{equation}
\label{weights_is}
\bG\leftarrow \bG\otimes\frac{\bB^T\frac{\bV}{\left(\bB\bG\right)^2}}{\bB^T\frac{\Bone}{\bB\bG}},	
\end{equation}
where the operation $\otimes$ is an element-wise multiplication, all divisions and $(.)^2$ are element-wise operations. In source separation applications, IS-NMF is used with matrices of power spectral densities of the source signals \citep{Fevotte:09:nmfwisd,xabier:11:assdnmfss}.  
     
\section{Problem formulation for SCSS}
\label{sp}
In single channel source separation (SCSS) problems, the aim is to find estimates of source signals that are mixed on a single observation channel $y(t)$. This problem is usually solved in the short time Fourier transform (STFT) domain. Let $Y(t,f)$ be the STFT of $y(t)$, where $t$ represents the frame index and $f$ is the frequency-index. Due to the linearity of the STFT, we have:   
\begin{equation}
\label{stftsum}
Y(t,f)=S_1(t,f)+S_2(t,f),
\end{equation}
where $S_1(t,f)$ and $S_2(t,f)$ are the unknown STFT of the first and second sources in the mixed signal. Assuming independence of the sources, we can write the power spectral density (PSD) of the measured signal as the sum of source signal PSDs as follows: 
\begin{equation}
	\sigma^{2}_{y}(t,f)=\sigma_1^{2}(t,f)+\sigma_2^{2}(t,f),
\end{equation}
where $\sigma^2_y(t,f)=E(|Y(t,f)|^2)$. We can write the PSDs for all frames as a spectrogram matrix as follows:
\begin{equation}
\label{unkowns}
	\bY=\bS_1+\bS_2, 
\end{equation}
where $\bS_1$ and $\bS_2$ are the unknown spectrograms of the source signals, and they need to be estimated using the observed mixed signal and training data for each source. The spectrogram of the measured signal $\bY$ is calculated by taking the squared magnitude of the STFT of the measured signal $y(t)$. 

\section{Conventional NMF for SCSS}
\label{conv_nmf}
In conventional single channel source separation using NMF without regularization \citep{emad:12:avsrwbmscss}, there are two main stages to find estimates for $\bS_1$ and $\bS_2$ in equation (\ref{unkowns}). The first stage is the training stage and the second stage is the separation/testing stage. In the training stage, the spectrogram $\bS^{\scriptsize{\mbox{train}}}$ for each source is calculated by computing the squared magnitude of the STFT of each source training signal. NMF is used to decompose the spectrogram into basis and gains matrices as follows:
\begin{equation}
\label{train_source}
\bS_1^{\scriptsize{\mbox{train}}}\approx\bB_1\bG_1^{\scriptsize{\mbox{train}}}, \ \ \ \ \ \ \ \bS_2^{\scriptsize{\mbox{train}}}\approx\bB_2\bG_2^{\scriptsize{\mbox{train}}},	
\end{equation}
the multiplicative update rules in equations (\ref{basis_is}) and (\ref{weights_is}) are used to solve for $\bB_1, \bB_2, \bG_1^{\scriptsize{\mbox{train}}}$ and $\bG_2^{\scriptsize{\mbox{train}}}$ for both sources. Within each iteration, the columns of $\bB_1$ and $\bB_2$ are normalized and the matrices $\bG_1^{\scriptsize{\mbox{train}}}$ and $\bG_2^{\scriptsize{\mbox{train}}}$ are computed accordingly. The initialization of all matrices $\bB_1, \bB_2, \bG_1^{\scriptsize{\mbox{train}}}$ and $\bG_2^{\scriptsize{\mbox{train}}}$ is done using positive random noise. After finding basis and gains matrices for each source training data, the basis matrices are used in the mixed signal decomposition as shown in the following sections. All the basis matrices $\bB_1$ and $\bB_2$ are kept fixed in the remaining sections in this paper.

In the separation stage after observing the mixed signal $y(t)$, NMF is used to decompose the mixed signal spectrogram $\bY$ with the trained bases matrices $\bB_1$ and $\bB_2$ that were found from solving equation (\ref{train_source}) as follows:
\begin{equation}
\label{decomp}
\bY\approx
\left[
\bB_1,\bB_2
\right]\bG, \ \ \ \mbox{or} \ \ \ 
\bY\approx
\left[
\bB_1 \ \ \ \bB_2
\right]\left[
\begin{array}[pos]{c}
\bG_1 \\
\bG_2
\end{array}	
\right],
\end{equation}
then the corresponding spectrogram estimate for each source can be found as:
\begin{equation}
\label{sest1}
\widetilde{\bS}_1=\bB_1\bG_1, \ \ \ \ \ \widetilde{\bS}_2=\bB_2\bG_2.	
\end{equation}
Let $\bB_{train}=\left[\bB_1,\bB_2\right]$. The only unknown here is the gains matrix $\bG$ since the matrix $\bB_{train}$ was found during the training stage and it is fixed in the separation stage. The matrix $\bG$ is a combination of two submatrices as in equation (\ref{decomp}). NMF is used to solve for $\bG$ in (\ref{decomp}) using the update rule in equation (\ref{weights_is}) and $\bG$ is initialized with positive random numbers. 

\section{Motivation for regularized NMF}
\label{motivation_RNMF}
The solution of the gains submatrix $\bG_1$ in (\ref{decomp}) is affected by the existence of the second source in the mixed signal. Also, $\bG_2$ is affected by the first source in the mixed signal. The effect of one source into the gains matrix solution of the other source strongly depends on the energy level of each source in the mixed signal. Therefore, the estimated spectrograms $\widetilde{\bS}_1$ and $\widetilde{\bS}_2$ in equation (\ref{sest1}) that are found from solving $\bG$ using the update rule in (\ref{weights_is}) may contain residual contribution from each other and other distortions. To fix this problem, more discriminative constraints must be added to the solution of each gains submatrix. The columns of the solution gains submatrix $\bG_1$ and $\bG_2$ should form a valid/expected weight combinations for its corresponding basis matrix of its corresponding source signal. The information about the valid weight combinations that can exits in the gains matrix for a source signal can be found in the gains matrix that was computed from the clean training data of the same source. For example, the information about valid weight combinations that can exist in the gains matrix $\bG_1$ in equation (\ref{decomp}) can be found in its training gains matrix $\bG_1^{\scriptsize{\mbox{train}}}$ in equation (\ref{train_source}). The columns of the trained gains matrix $\bG_1^{\scriptsize{\mbox{train}}}$ represent the valid weight combinations that the basis matrix $\bB_1$ can receive for the first source. Note that, the basis matrix $\bB_1$ is common in the training and separation stages. The solution of the gains submatrix $\bG_1$ in equation (\ref{decomp}) should consider the prior about the valid combination that is present in its corresponding trained gains matrix $\bG_1^{\scriptsize{\mbox{train}}}$ in equation (\ref{train_source}) for the same source. 

In our previous work \citep{grais:12:rnmfgmpsscss}, data in the training gains matrix $\bG_i^{\scriptsize{\mbox{train}}}$ for source $i$ was modeled using a GMM. The NMF solution of the gains matrix during the separation stage was guided by the prior GMM. The GMM was learned using the logarithm of the normalized columns of the training gains matrix.  
The NMF solution for the gains matrix during the separation stage was enforced to increase its log-likelihood with the trained GMM prior using regularized NMF as follows:
\begin{equation}
\label{div_sep21}
C_{old}=D_{IS}\left(\bY\left|\right|\bB_{train}\bG\right)- R_{old}(\bG),
\end{equation}
where $R_{old}(\bG)$ is the weighted sum of the log-likelihoods of the log-normalized columns of the gains matrix $\bG$. $R_{old}(\bG)$ was defined as follows:
\begin{equation}
	R_{old}(\bG)=\sum_{i=1}^2\eta_i\Gamma_{old}(\bG_i), 
\end{equation}
where $\Gamma_{old}(\bG_i)$ is the log-likelihood for the submatrix $\bG_i$, and $\eta_i$ is the regularization parameter for source $i$. The regularization parameter in \citet{grais:12:rnmfgmpsscss} was playing two important roles. The first role was to match the scale of the IS-NMF divergence term with the scale of the log-likelihood prior term. The second role was to decide how much the regularized NMF cost function needs to rely on the prior term. The results in \citet{grais:12:rnmfgmpsscss} show that, when the source $i$ has higher energy level than the other sources, the value of its corresponding regularization parameter $\eta_i$ becomes smaller than the values of other regularization parameters for the other sources. That can be reformed as follows: when the source has high energy level, the gains matrix solution of the regularized NMF in (\ref{div_sep21}) rely less on the prior model and vice versa. The values of the regularization parameters in \citet{grais:12:rnmfgmpsscss} was chosen manually for every energy level for each source. In the cases when the conjugate prior models of the NMF solutions were used \citep{virtanen:08:betnnffasm,canny:04:gap}, the hyper-parameters of the prior models were also chosen manually. The conjugate prior models usually enforced on NMF solutions using a Bayesian framework \citep{Fevotte:09:nmfwisd,virtanen:08:betnnffasm,canny:04:gap}. In \citet{grais:12:rnmfgmpsscss}, it was also shown that, the hyper-parameter choices for the conjugate prior models can also depend on the energy level differences of the source signals in the mixed signal.     

\section{Motivation for the proposed regularized NMF}
\label{motivation_RNMF2}
In this work, we try to use prior GMMs to guide the solution of the gains matrix during the separation stage using regularized NMF as in \citet{grais:12:rnmfgmpsscss} but following a totally different regularization strategy. We also try to find a way to estimate how much the solution of the regularized NMF needs to rely on the prior GMMs automatically not manually as in \citet{grais:12:rnmfgmpsscss}. The way of finding how much the regularized NMF solution of the gains matrix needs to rely on the prior GMM is by measuring how far the statistics of the solution of the gains matrix $\bG_i$ in (\ref{decomp}) is from the statistics of the solution of the valid gains matrix solution $\bG_i^{\scriptsize{\mbox{train}}}$ in (\ref{train_source}) for source $i$. Recall that, the matrix $\bG_i^{\scriptsize{\mbox{train}}}$ in (\ref{train_source}) contains the weight combinations that the columns in the basis matrix $\bB_i$ can jointly receive for the clean data of source $i$. The data in $\bG_i^{\scriptsize{\mbox{train}}}$ can be used as a prior information for what kinds of weight combinations that should exist in $\bG_i$ in (\ref{decomp}) since the matrix $\bB_i$ is the same in (\ref{train_source}) and (\ref{decomp}). The matrix $\bG_i^{\scriptsize{\mbox{train}}}$ in (\ref{train_source}) is used to train a prior GMM for the expected (valid) weight combinations that can exist in the gains matrix for source $i$ as in \citet{grais:12:rnmfgmpsscss}. The solution of the gains submatrix $\bG_i$ in (\ref{decomp}) can be seen as a deformed observation that needs to be restored using MMSE estimate under its corresponding GMM prior for source $i$. How far the statistics of the solution of the gains matrix $\bG_i$ is from the statistics of the solution of the valid gains matrix solution $\bG_i^{\scriptsize{\mbox{train}}}$ can be seen as how much the gains submatrix $\bG_i$ is deformed. How much deformation exists in the gains matrix $\bG_i$ can be learned directly and the logarithm of this deformation is modeled using a Gaussian distribution with zero mean and a diagonal covariance matrix $\bPsi_i$. When the deformation or the uncertainty measurement $\bPsi_i$ of the gain submatrix $\bG_i$ is high, we expect our target regularized NMF cost function to rely more on the prior GMM for source $i$ and vice versa. Based on the measurement $\bPsi_i$, the proposed NMF cost function decides automatically how much the solution of the regularized NMF needs to rely on the prior GMMs, which is a main advantage of the proposed regularized NMF over our previous work \citep{grais:12:rnmfgmpsscss}. Applying the prior information on the gains matrix $\bG_i$ in (\ref{decomp}) using MMSE estimate under a GMM prior using regularized NMF is the new strategy that we introduce in this paper. 

In the following sections, we give more details about training the prior GMM for the gains matrix for each source. Then, we give more details about our proposed regularized NMF using MMSE estimate to find better solution for the gains matrix in (\ref{decomp}). In Section \ref{CNMF}, we present our proposed regularized NMF in a general manner. In Section \ref{CNMF}, we assume we have a trained basis matrix $\bB$, a trained prior GMM for a clean gains matrix, and a gains matrix $\bG$ that inherited some distortion from the original matrix $\bV$ from solving equation (\ref{div_is}). We introduce our proposed regularized NMF in a general fashion in Section \ref{CNMF} to make the idea clearer for different NMF applications like, dimensionality reduction, denosing, and other applications. The update rules that solve the proposed regularized NMF are also derived in Section \ref{CNMF} in a general fashion regardless of the application. The GMM in Section \ref{CNMF} is the trained prior GMM that captures the statistics of the valid weights combinations that should have been existed in the gains matrix $\bG$. In section \ref{r_CNMF}, we show how we use the proposed regularized NMF to find better solutions for the gain submatrices in equation (\ref{decomp}) for our single channel source separation problem.      
%

\section{Training the GMM prior models}
\label{training_prior}                      
We use the gains matrices $\bG_1^{\scriptsize{\mbox{train}}}$ and $\bG_2^{\scriptsize{\mbox{train}}}$ in equation (\ref{train_source}) to train prior models for the expected/valid weight patterns in the gains matrix for each source. For each matrix $\bG_1^{\scriptsize{\mbox{train}}}$ and $\bG_2^{\scriptsize{\mbox{train}}}$, we normalize their columns and then calculate their logarithm. The normalization in this paper is done using the Euclidean norm. The log-normalized columns are then used to train a gains prior GMM for each source. The GMM for a random variable $\Bx$ is defined as:
\begin{equation}
\label{GMM}
p(\Bx)=\sum_{k=1}^{K}\frac{\pi_k}{(2\pi)^{d/2}\left|\bSigma_k\right|^{1/2}}\exp\left\{-\frac{1}{2}\left(\Bx-\Bmu_k\right)^T\bSigma_k^{-1}\left(\Bx-\Bmu_k\right)\right\},
\end{equation}
where $K$ is the number of Gaussian mixture components, $\pi_k$ is the mixture weight, $d$ is the vector dimension, $\Bmu_k$ is the mean vector and $\bSigma_k$ is the diagonal covariance matrix of the $k^{th}$ Gaussian model. In training GMM, the expectation maximization (EM) algorithm \citep{Dempster:77:mlidvema} is used to learn the GMM parameters ($\pi_k, \Bmu_k, \bSigma_k, \ \ \forall k=\left\{1,2,...,K\right\}$) for each source given its trained gain matrix $\bG^{\scriptsize{\mbox{train}}}$. The suitable value for $K$ usually depends on the nature, dimension and the size of the available training data. We use the logarithm because it has been shown that the logarithm of a variable taking values between 0 and 1 can be modeled well by a GMM \citep{wessel:00:upw}. 
Since the main goal of the prior model is to capture the statistics of the patterns in the trained gains matrix, we use normalization to make the prior models insensitive to the energy level of the training data. The normalization makes the same prior models applicable for a wide range of energy levels and avoids the need to train a different prior model for different energy levels. By normalization we are modeling the ratio and correlation between the combination of the weights that the bases can jointly receive. 

%
\section{The proposed regularized NMF}
\label{CNMF}
The goal of regularized NMF is to incorporate prior information on the solution matrices $\bB$ and $\bG$. In this work, we enforce a statistical prior information on the solution of the gains/weights matrix $\bG$ only. We need the solution of the gains matrix $\bG$ to minimize the IS-divergence cost function in equation (\ref{div_is}), and the columns of the gains matrix $\bG$ should form valid weight combinations under a prior GMM model. 

The most used strategy for incorporating a prior is by maximizing the likelihood of the solution under the prior model while minimizing the NMF divergence at the same time. To achieve this, we usually add these two objectives in a single cost function. In \citet{grais:12:rnmfgmpsscss}, a GMM was used as the prior model for the gains matrix, and the solution of the gains matrix was encouraged to increase its log-likelihood with the prior model using this regularized NMF cost function. The regularization parameters in \citet{grais:12:rnmfgmpsscss} were the only tools to control how much the regularized NMF relies on the prior models based on the energy differences of the sources in the mixed signal. The values of the regularization parameters were changed manually in that work. 

Gaussian mixture model is a very general prior model where we can see the means of the GMM mixture components as ``valid templates'' that were observed in the training data. Even, Parzen density priors \citep{Kim:07:nspfacbýs} can be seen under the same framework. In Parzen density prior estimation, training examples are seen as ``valid templates'' and a fixed variance is assigned to each example. In GMM priors, we learn the templates as cluster means from training data and we can also estimate the cluster variances from the data. We can think of the GMM prior as a way to encourage the use of valid templates or cluster means in the NMF solution during the separation stage. This view of the GMM prior will be helpful in understanding the MMSE estimate method we introduce in this paper.

%
We can find a way of measuring how far the conventional NMF (NMF without prior) solution is from the trained templates in the prior GMM and call this the error term. Based on this error, the regularized NMF can decide automatically how much the solution of the NMF needs help from the prior model. If the conventional NMF solution is far from the templates then the regularized NMF will rely more on the prior model. If the conventional NMF solution is close to the templates then the regularized NMF will rely less on the prior model. By deciding automatically how much the regularized NMF needs to rely on the prior we conjecture that, we do not need to manually change the values for the regularization parameter based on the energy differences of the sources in the mixed signal \footnote{In this paper, the regularization parameters are chosen once and kept fixed regardless of the energy differences of the source signals} to improve the performance of NMF as in \citet{grais:12:rnmfgmpsscss}. 
   
We use the following way of measuring how far the conventional NMF solution is from the prior templates:
We can see the solution of the conventional NMF as distorted observations of a true/valid template. Given the prior GMM templates, we can learn a probability distribution model for the distortion that captures how far the observations in the conventional gains matrix is from the prior GMM. The distortion or the error model can be seen as a summary of the distortion that exists in all columns in the gains matrix of the NMF solution.

Based on the prior GMM and the trained distortion model, we can find a better estimate for the desired observation for each column in the distorted gains matrix. We can mathematically formulate this by seeing the solution matrix $\bG$ that only minimizes the cost function in equation (\ref{div_is}) as a distorted image where its restored image needs to be estimated. The columns of the matrix $\bG$ are normalized using the $\ell^2$ norm and their logarithm is then calculated. Let the log-normalized column $n$ namely ($\log\frac{\Bg_n}{\left\|\Bg_n\right\|_2}$) of the gains matrix be $\Bq_n$. The vector $\Bq_n$ is treated as a distorted observation as:
\begin{equation}    
\label{corrupted1}
\Bq_n=\Bx_n+\Be,
\end{equation} 
where $\Bx_n$ is the logarithm of the unknown desired pattern that corresponds to the observation $\Bq_n$ and needs to be estimated under a prior GMM, $\Be$ is the logarithm of the deformation operator, which is modeled by a Gaussian distribution with zero mean and diagonal covariance matrix $\bPsi$ as $\mbox{\rsfsfive{N}}\left(\Be|\Bzero,\bPsi \right)$. The GMM prior model for the gains matrix is trained using log-normalized columns of the trained gains matrix from training data as shown for example in Section \ref{training_prior}. The uncertainty $\bPsi$ is trained directly from all the log-normalized columns of the gains matrix $\Bq=\left\{\Bq_1, .., \Bq_n, .., \Bq_N\right\}$, where $N$ is the number of columns in the matrix $\bG$. The uncertainty $\bPsi$ can be iteratively learned using the expectation maximization (EM) algorithm. Given the prior GMM parameters which are considered fixed here, the update of $\bPsi$ is found based on the sufficient statistics $\hat{\Bz}_n$ and $\hat{R}_n$ as follows \citep{rosti:01:glgm,rosti:04:fahmmfsr,zoubin:97:emmfa} [Appendix A]:
\begin{equation}
\label{epsi_estmate_train1}
\bPsi=\mbox{diag} \left\{ \frac{1}{N}\sum^N_{n=1}\left(\Bq_n\Bq_n^T-\Bq_n\hat{\Bz}_n^T-\hat{\Bz}_n\Bq_n^T+\hat{\bR}_n\right)   \right\},
\end{equation}
where the ``diag'' operator sets all the off-diagonal elements of a matrix to zero, $N$ is the number of columns in matrix $\bG$, and the sufficient statistics $\hat{\Bz}_n$ and $\hat{R}_n$ can be updated using $\bPsi$ from the previous iteration as follows: 
\begin{equation}
\label{gool_expl2}
\hat{\Bz}_n=\sum_{k=1}^K\gamma_{kn}\hat{\Bz}_{kn},
\end{equation}
and
\begin{equation}
\label{cond_corr_train_final}
\hat{\bR}_n=\sum_{k=1}^K \gamma_{kn}\hat{\bR}_{kn},
\end{equation}
where
\begin{equation}
\gamma_{kn}=\left[\frac{\pi_k \mbox{\rsfsfive{N}} \left(\Bq_n|\Bmu_k,\bSigma_k+\bPsi \right)}{\sum_{j=1}^K \pi_j \mbox{\rsfsfive{N}} \left(\Bq_n|\Bmu_j,\bSigma_j+\bPsi \right)}\right],
\end{equation}
\begin{equation}
\label{cond_corr_train}
\hat{\bR}_{kn}=\bSigma_k-\bSigma_k\left(\bSigma_k+\bPsi\right)^{-1}\bSigma_k^T+\hat{\Bz}_{kn}\hat{\Bz}_{kn}^T,
\end{equation} 
and
\begin{equation}
\label{cond_expect_train1}
\hat{\Bz}_{kn}=\Bmu_k+\bSigma_k\left(\bSigma_k+\bPsi\right)^{-1}\left(\Bq_n-\Bmu_k\right).
\end{equation}  
$\bPsi$ is considered as a general uncertainty measurement over all the observations in matrix $\bG$. $\bPsi$ can be seen as a model that summarizes the deformation that exists in all columns in the gains matrix $\bG$.

Given the GMM prior parameters and the uncertainty measurement $\bPsi$, the MMSE estimate of each pattern $\Bx_n$ given its observation $\Bq_n$ under the observation model in equation (\ref{corrupted1}) can be found similar to \citet{rosti:01:glgm,rosti:04:fahmmfsr}, and \citet{zoubin:97:emmfa} as in Appendix A as follows: 
\begin{equation}
\label{gool_expl}
f\left(\Bq_n \right)=\sum_{k=1}^K\gamma_{kn}\left[\Bmu_k+\bSigma_k\left(\bSigma_k+\bPsi\right)^{-1}\left(\Bq_n-\Bmu_k\right)\right]=\hat{\Bx}_n,
\end{equation}
where
\begin{equation}
\gamma_{kn}=\left[\frac{\pi_k \mbox{\rsfsfive{N}} \left(\Bq_n|\Bmu_k,\bSigma_k+\bPsi \right)}{\sum_{j=1}^K \pi_j \mbox{\rsfsfive{N}} \left(\Bq_n|\Bmu_j,\bSigma_j+\bPsi \right)}\right].
\end{equation}
The value of $\bPsi$ in the term $\bSigma_k\left(\bSigma_k+\bPsi\right)^{-1}$ in equation (\ref{gool_expl}) plays an important role in this framework.  
When the entries of the uncertainty $\bPsi$ are very small comparing to their corresponding entries in $\bSigma_k$ for a certain active GMM component $k$, the term $\bSigma_k\left(\bSigma_k+\bPsi\right)^{-1}$ tends to be the identity matrix, and MMSE estimate in (\ref{gool_expl}) will be the observation $\Bq_n$. When the entries of the uncertainty $\bPsi$ are very high comparing to their corresponding entries in $\bSigma_k$ for a certain active GMM component $k$, the term $\bSigma_k\left(\bSigma_k+\bPsi\right)^{-1}$ tends to be a zeros matrix, and MMSE estimate will be the weighted sum of prior templates $\sum_{k=1}^K\gamma_{kn}\Bmu_k$. In most cases $\gamma_{kn}$ tends to be close to one for one Gaussian component, and close to zero for the other components in a large dimension space. This makes the MMSE estimate in the case of high $\bPsi$ to be one of the mean vectors in the prior GMM, which is considered as a template pattern for the valid observation. We can rephrase this as follows: When the uncertainty of the observations $\Bq$ is high, the MMSE estimate of $\Bx$, relies more on the prior GMM of $\Bx$. When the uncertainty of the observations $\Bq$ is low, the MMSE estimate of $\Bx$, relies more on the observation $\Bq_n$. In general, the MMSE solution of $\Bx$ lies between the observation $\Bq_n$ and one of the templates in the prior GMM. The term $\bSigma_k\left(\bSigma_k+\bPsi\right)^{-1}$ controls the distance between $\hat{\Bx}_n$ and $\Bq_n$ and also the distance between $\hat{\Bx}_n$ and one of the template $\Bmu_k$ assuming that $\gamma_{kn}\approx 1$ for a Gaussian component $k$. 

The model in equation (\ref{corrupted1}) expresses the normalized columns of the gains matrix as a distorted image with a multiplicative deformation diagonal matrix. For the normalized gain columns $\frac{\Bg_n}{\left\|\Bg_n\right\|_2}$ of $\bG$ there is a deformation matrix $\bE$ with log-normal distribution that is applied to the correct pattern that we need to estimate $\hat{\Bg}_{n}$ as follows:
\begin{equation}    
\label{corrupted2}
\frac{\Bg_{n}}{\left\|\Bg_{n}\right\|_2}=\bE\hat{\Bg}_{n}.
\end{equation}  
The uncertainty for $\bE$ is represented in its covariance matrix $\bPsi$. 
For the distorted matrix $\bG$ we find its corresponding MMSE estimate for its log-normalized columns $\hat {\bG}$. Another reason for working in logarithm domain is that, the gains are constrained to be nonnegative and the MMSE estimate can be negative so the logarithm of the normalized gains is an unconstrained variable that we can work with. The estimated weight patterns in $\hat {\bG}$ that are corresponding to the MMSE estimates for the correct patterns do not consider minimizing the NMF cost function in equation (\ref{div_is}), which is still the main goal. We need the solution of $\bG$ to consider the pattern shape priors on the solution of the gains matrix, and also considers the reconstruction error of the NMF cost function. To consider the combination of the two objectives, we consider using the regularized NMF. We add a penalty term to the NMF-divergence cost function. The penalty term tries to minimize the distance between the solution of log-normalized columns of $\Bg_n$ with its corresponding MMSE estimate $f(\Bg_n)$ as follows:
\begin{equation}
\label{penalty}
\footnotesize{
\log\frac{\Bg_n}{\left\|\Bg_n\right\|_2}\approx f\left( \log \frac{\Bg_n}{\left\|\Bg_n\right\|_2} \right) \ \  \mbox{or} \ \ \  \frac{\Bg_n}{\left\|\Bg_n\right\|_2}\approx \exp \left(f\left( \log\frac{\Bg_n}{\left\|\Bg_n\right\|_2} \right) \right).
}
\end{equation}
The regularized IS-NMF cost function is defined as follows: 
\begin{equation}
\label{div_sep1}
C=D_{IS}\left(\bV\left|\right|\bB\bG\right)+ \alpha L(\bG),
\end{equation}   
where
\begin{equation}
\label{first_constr1}
L(\bG)=\sum_{n=1}^N \left\|\frac{\Bg_{n}}{\left\|\Bg_{n}\right\|_2}-\exp \left(f\left( \log\frac{\Bg_n}{\left\|\Bg_n\right\|_2} \right) \right)\right\|_2^2, 
\end{equation}
$f\left( \log\frac{\Bg_n}{\left\|\Bg_n\right\|_2} \right)$ is the MMSE estimate defined in equation (\ref{gool_expl}), and $\alpha$ is a regularization parameter. The regularized NMF can be rewritten in more details as
\begin{equation}
\label{div_sep_CONS}
\scriptsize{
C=\sum_{m,n}\left(\frac{\bV_{m,n}}{\left(\bB\bG\right)_{m,n}}-\log\frac{\bV_{m,n}}{\left(\bB\bG\right)_{m,n}}-1\right)+
\alpha\sum_{n=1}^N \left\|\frac{\Bg_{n}}{\left\|\Bg_{n}\right\|_2}-\exp\left( \sum_{k=1}^K\gamma_{kn}\left[\Bmu_k+\bSigma_k\left(\bSigma_k+\bPsi\right)^{-1}\left(\log\frac{\Bg_{n}}{\left\|\Bg_{n}\right\|_2}-\Bmu_k\right)\right] \right)\right\|_2^2.
}
\end{equation}   
In equation (\ref{div_sep_CONS}), the MMSE estimate of the desired patterns of the gains matrix is embedded in the regularized NMF cost function. The first term in (\ref{div_sep_CONS}), decreases the reconstruction error between $\bV$ and $\bB\bG$. Given $\bPsi$, we can forget for a while the MMSE estimate concept that leaded us to our target regularized NMF cost function in (\ref{div_sep_CONS}) and see equation (\ref{div_sep_CONS}) as an optimization problem. We can see from (\ref{div_sep_CONS}) that, if the distortion measurement parameter $\bPsi$ is high, the regularized nonnegative matrix factorization solution for the gains matrix will rely more on the prior GMM for the gains matrix. If the distortion parameter $\bPsi$ is low, the regularized nonnegative matrix factorization solution for the gains matrix will be close to the ordinary NMF solution for the gains matrix without considering any prior. The second term in equation (\ref{div_sep_CONS}) is ignored in the case of zero uncertainty $\bPsi$. In case of high values of $\bPsi$, the second term encourages to decrease the distance between each normalized column $\frac{\Bg_{n}}{\left\|\Bg_{n}\right\|_2}$ in $\bG$ with a corresponding prior template $\exp\left(\Bmu_k\right)$ assuming that $\gamma_{kn}\approx 1$ for a certain Gaussian component $k$. For different values $\bPsi$, the penalty term decreases the distance between each $\frac{\Bg_{n}}{\left\|\Bg_{n}\right\|_2}$ and an estimated pattern that lies between a prior template and $\frac{\Bg_{n}}{\left\|\Bg_{n}\right\|_2}$. The term ($\log\frac{\Bg_{n}}{\left\|\Bg_{n}\right\|_2}-\Bmu_k$) in (\ref{div_sep_CONS}) measures how far each log-normalized column in the gains matrix is from a valid template $\Bmu_k$. Under the assumption $\gamma_{kn}\approx 1$ for a certain Gaussian component $k$, the second term in (\ref{div_sep_CONS})  is also ignored when the observation $\log\frac{\Bg_{n}}{\left\|\Bg_{n}\right\|_2}$ form a valid pattern ($\log\frac{\Bg_{n}}{\left\|\Bg_{n}\right\|_2}=\Bmu_k$). How far each log-normalized column in the gains matrix is from a valid template decides how much influence the MMSE estimate prior term has to the solution of (\ref{div_sep_CONS}) for each observation.     

The multiplicative update rule for $\bB$ in (\ref{div_sep_CONS}) is still the same as in equation (\ref{basis_is}). The multiplicative update rule for $\bG$ can be found by following the same procedures as in~\citet{Virtanen:07:msssbnmfwtcasc,nancy:10:ehsbnmf,grais:12:rnmfgmpsscss}. The gradient with respect to $\bG$ of the cost function $\nabla_G C$ can be expressed as a difference of two positive terms $\nabla^+_G C$ and $\nabla^-_G C$ as follows:     
\begin{equation}
\label{div_grad}
\nabla_G C=	\nabla^+_G C - \nabla^-_G C.
\end{equation}
The cost function is shown to be nonincreasing under the update rule~\citep{Virtanen:07:msssbnmfwtcasc, nancy:10:ehsbnmf}:
\begin{equation}
\label{grad_updateG}
\bG\leftarrow \bG\otimes\frac{\nabla^-_G C}{\nabla^+_G C}, 
\end{equation}
where the operations $\otimes$ and division are element-wise as in equation (\ref{weights_is}). We can write the gradients as:
\begin{equation}
\label{total_gradg}
\nabla_G C=\nabla_G D_{IS} + \alpha \nabla_G L(\bG),
\end{equation}
where $\nabla_G L(\bG)$ is a matrix with the same size of $\bG$. The gradient for the IS-NMF and the gradient of the prior term can also be expressed as a difference of two positive terms as follows: 
\begin{equation}
\label{total_grad2}
\nabla_G D_{IS}=\nabla_G^+ D_{IS}-\nabla_G^- D_{IS},
\end{equation}
and
\begin{equation}
\label{total_gradL}
\nabla_G L(\bG)=\nabla_G^+L(\bG) -\nabla_G^-L(\bG).
\end{equation}
We can rewrite equations (\ref{div_grad}, \ref{total_gradg}) as:
\begin{equation}
\label{total_gradg_detail}
\nabla_G C=\left( \nabla_G^+ D_{IS} + \alpha \nabla_G^+L(\bG) \right) - \left( \nabla_G^- D_{IS} + \alpha \nabla_G^-L(\bG)  \right).
\end{equation}
The final update rule in equation (\ref{grad_updateG}) can be written as follows:
\begin{equation}
\label{grad_updateG_final}
\bG\leftarrow \bG\otimes\frac{\nabla_G^- D_{IS} + \alpha \nabla_G^-L(\bG)}{\nabla_G^+ D_{IS} + \alpha \nabla_G^+L(\bG)}, 
\end{equation}
where
\begin{equation}
\label{total_grad_is}
\nabla_G D_{IS}=\bB^T\frac{\Bone}{\bB\bG}-\bB^T\frac{\bV}{\left(\bB\bG\right)^2},
\end{equation}

\begin{equation}
\label{total_grad_neg_is}
\nabla_G^- D_{IS}=\bB^T\frac{\bV}{\left(\bB\bG\right)^2}, \ \ \ \ \ \mbox{and} \ \ \ \ \  \nabla_G^+ D_{IS}=\bB^T\frac{\Bone}{\bB\bG}.
\end{equation}
Note that, in calculating the gradients $\nabla_G^+L(\bG)$ and $\nabla_G^-L(\bG)$, the term $\gamma_{kn}$ is also a function of $\bG$. The gradients $\nabla_G^+L(\bG)$ and $\nabla_G^-L(\bG)$ are calculated in Appendix B. Since all the terms in equation (\ref{grad_updateG_final}) are nonnegative, then the values of $\bG$ of the update rule (\ref{grad_updateG_final}) are nonnegative. 

%
\section{The proposed regularized NMF for SCSS}
\label{r_CNMF}
\begin{figure}[!htb]
  \centering
   \centerline{\includegraphics[width=1\textwidth]{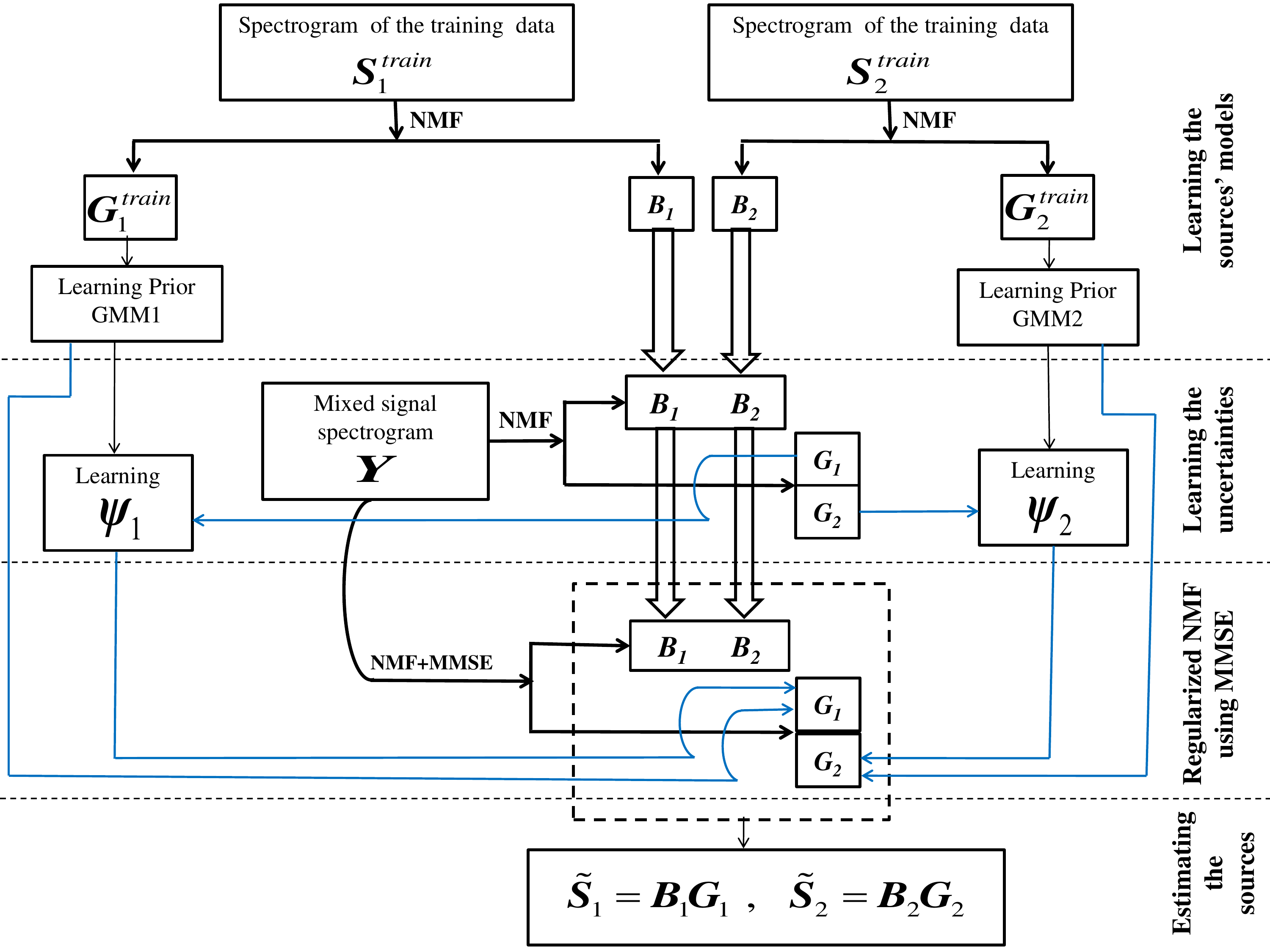}}
\caption{The flow chart of using regularized NMF with MMSE estimates under GMM priors for SCSS. The term NMF+MMSE means regularized NMF using MMSE estimates under GMM priors.}
\label{flow}
\end{figure}

In this section, we are back to the single channel source separation problem to find a better solution to equation (\ref{decomp}). Figure \ref{flow} shows the flow chart that summarizes all stages of applying our proposed regularized NMF method for SCSS problems. 
Given the trained basis matrices $\bB_1$, $\bB_2$ that were computed from solving (\ref{train_source}), and the trained gain prior GMM for each source from Section \ref{training_prior}, we try to apply the proposed regularized NMF cost function in Section \ref{CNMF} to find better solution for the gain submatrices in equation (\ref{decomp}). The bases matrix $\bB_{train}=\left[\bB_1,\bB_2\right]$ is still fixed here, we just need to update the gains matrix $\bG$ in (\ref{decomp}). The normalized columns of the submatrices $\bG_1$ and $\bG_2$ in equation (\ref{decomp}) can be seen as deformed images as in equation (\ref{corrupted2}) and their restored images are needed to be estimated. First, we need to learn the uncertainties parameters $\bPsi_1$ and $\bPsi_2$ for the deformation operators $\bE_1$ and $\bE_2$ respectively for each image as shown in learning the uncertainties stage in Figure \ref{flow}. The columns of the submatrix $\bG_1$ are normalized and their logarithm are calculated and used with the trained GMM prior parameters for the first source to estimate $\bPsi_1$ iteratively using the EM algorithm in equations (\ref{epsi_estmate_train1}) to (\ref{cond_expect_train1}). The log-normalized columns ``$\log\frac{\Bg_n}{\left\|\Bg_n\right\|_2}$'' of $\bG_1$ can be seen as $\Bq_n$ in equations (\ref{epsi_estmate_train1}) to (\ref{cond_expect_train1}). We repeat the same procedures to calculate $\bPsi_2$ using the log-normalized columns of $\bG_2$ and the prior GMM for the second source. The uncertainties $\bPsi_1$ and $\bPsi_2$ can also be seen as measurements of the remaining distortion from one source into another source, which also depends on the mixing ratio between the two sources. For example, if the first source has higher energy than the second source in the mixed signal, we expect the values of $\bPsi_2$ to be higher than the values in $\bPsi_1$ and vice versa. After calculating the uncertainty parameters for both sources $\bPsi_1$ and $\bPsi_2$, we use the regularized NMF in (\ref{div_sep1}) to solve for $\bG$ with the prior GMMs for both sources and the estimated uncertainties $\bPsi_1$ and $\bPsi_2$ as follows:                    
\begin{equation}
\label{div_sep_source_sep}
C=D_{IS}\left(\bY\left|\right|\bB_{train}\bG\right)+ R(\bG),
\end{equation}   
where
\begin{equation}
\label{R}
R(\bG)=\alpha_1 L_1(\bG_1)+\alpha_2 L_2(\bG_2),
\end{equation}   
$L_1(\bG_1)$ is defined as in equation (\ref{first_constr1}) for the first source, $L_2(\bG_2)$ is for the second source,  $\alpha_1$, and $\alpha_2$ are their corresponding regularization parameters. The update rule in equation (\ref{grad_updateG_final}) can be used to solve for $\bG$ after modifying it as follows:
\begin{equation}
\label{grad_updateG_final2}
\bG\leftarrow \bG\otimes\frac{\nabla_G^- D_{IS} + \nabla_G^-R(\bG)}{\nabla_G^+ D_{IS} + \nabla_G^+ R(\bG)},  
\end{equation} 
where $\nabla_G^+ R(\bG)$ and $\nabla_G^- R(\bG)$ are nonnegative matrices with the same size of $\bG$ and they are combinations of two submatrices as follows: 
\begin{equation}
\label{Lcompin}
\nabla_G^- R(\bG)=
\left[
\begin{array}[pos]{c}
\alpha_1 \nabla_G^- L(\bG_1) \\
\alpha_2 \nabla_G^- L(\bG_2) \\
\end{array}	
\right], \ \ \ \nabla_G^+ R(\bG)=
\left[
\begin{array}[pos]{c}
\alpha_1 \nabla_G^+ L(\bG_1) \\
\alpha_2 \nabla_G^+ L(\bG_2) \\
\end{array}	
\right],
\end{equation}
where $\nabla_G^+ L(\bG_1), \nabla_G^- L(\bG_1), \nabla_G^+ L(\bG_2)$, and $\nabla_G^- L(\bG_2)$ are calculated as in section \ref{CNMF} for each source. 

The normalization of the columns of the gain matrices are used in the prior term $R(\bG)$ and its gradient terms only. The general solution for the gains matrix of equation (\ref{div_sep_source_sep}) at each iteration is not normalized. The normalization is done only in the prior term since the prior models have been trained by normalized data before. Normalization is also useful in cases where the source signals occur with different energy levels from each other in the mixed signal. Normalizing the training and testing gain matrices gives the prior models a chance to work with any energy level that the source signals can take in the mixed signal regardless of the energy levels of the training signals. 

The regularization parameters in (\ref{R}) have only one role. They are chosen to match the scale between the NMF divergence term and the MMSE estimate prior term in the regularized NMF cost function in (\ref{div_sep_source_sep}). There is no need to change the values of the regularization parameters according to the energy differences of the source signals in the mixed signal as in \citet{grais:12:rnmfgmpsscss}. Reasonable values for the regularization parameters are chosen manually and kept fixed in this work. Another main difference between the regularized NMF in \citet{grais:12:rnmfgmpsscss} that is shown in equation (\ref{div_sep21}) and the proposed regularized NMF in this paper is related to the training procedures for the source models. In both works, the main aim of the training stage is to train the basis matrices and the gains prior GMMs for the source signals. In \citet{grais:12:rnmfgmpsscss}, to match between the way the trained models were used during training with the way they were used during separation, the basis matrices and the prior GMM parameters were learned jointly using the regularized NMF cost function in (\ref{div_sep21}). The joint training for the source models was introduced in \citet{grais:12:rnmfgmpsscss} to improve the separation performance. In joint training, after updating the gains matrix at each NMF iteration using the gain update rule for the regularized NMF in (\ref{div_sep21}),  the GMM parameters were then updated (retrained). Since, we needed to update (retrain) the GMM parameters at each NMF iteration, joint training slowed down the training of the source models in \citet{grais:12:rnmfgmpsscss}. Another problem of using joint training is that, we had other regularization parameters during the training stage that needed to be chosen. Using joint training duplicates the number of the regularization parameters that need to be chosen. Choosing the regularization parameters in \citet{grais:12:rnmfgmpsscss} was done using validation data. That means, in \citet{grais:12:rnmfgmpsscss} we had to train many source models (basis matrix and prior GMM) for different regularization parameter values. Then, we chose the best combination for the regularization parameter values in training and separation stages that gave the best results during the separation stage. In the case of using MMSE estimate regularization for NMF, we do not need to use joint training. In this paper, we do not need to consider solving the regularized NMF in (\ref{div_sep_CONS}) during the training stage to solve (\ref{train_source}). In the training stage, the training data for each source is assumed to be clean data. Since the spectrogram of each source training data represents clean source data, the NMF solution for the gains matrix can not be seen as a distorted image. Therefore, the deformation measurement parameter $\bPsi^{train}$ is a matrix of zeros. When $\bPsi^{train}=\Bzero$, the MMSE estimates prior term in (\ref{div_sep_CONS}) will disappear because $\sum_{k=1}^K\gamma_{kn}=1$. Then, the regularized NMF (\ref{div_sep_CONS}) becomes just NMF. That means, we do not need to use the regularized NMF during the training stage which is not the case in \citet{grais:12:rnmfgmpsscss}. Here in the training stage, we just need to use IS-NMF to decompose the spectrogram of the training data into trained basis and gains matrices. After the trained gains matrix is computed, it is used to train the prior GMM as shown in Sections \ref{conv_nmf} and \ref{training_prior}.           

\section{Source signals reconstruction}
\label{masks}
After finding the suitable solution for the gains matrix $\bG$ in Section \ref{r_CNMF}, the initial estimated spectrograms $\widetilde{\bS}_1$ and $\widetilde{\bS}_2$ can be calculated from (\ref{sest1}) and then used to build spectral masks as follows:
\begin{equation}
\label{mask}
\bH_1=\frac{\widetilde{\bS}_1}{\widetilde{\bS}_1+\widetilde{\bS}_2},  \ \ \ \ \ \ \  \bH_2=\frac{\widetilde{\bS}_2}{\widetilde{\bS}_1+\widetilde{\bS}_2},
\end{equation}
where the divisions are done element-wise. The final estimate of each source STFT can be obtained as follows:
\begin{equation}
\label{eq:deq5}
\hat {S_1}\left(t,f\right)=\bH_1\left(t,f\right)Y\left(t,f\right), \ \ \ \ \ \hat {S_2}\left(t,f\right)=\bH_2\left(t,f\right)Y\left(t,f\right),
\end{equation}
where $Y\left(t,f\right)$ is the STFT of the observed mixed signal in equation (\ref{stftsum}), $\bH_1\left(t,f\right)$ and $\bH_2\left(t,f\right)$ are the entries at row $f$ and column $t$ of the spectral masks $\bH_1$ and $\bH_2$ respectively. The spectral mask entries scale the observed mixed signal STFT entries according to the contribution of each source in the mixed signal. The spectral masks can be seen as the Wiener filter as in~\cite{Fevotte:09:nmfwisd}. The estimated source signals $\hat{s}_1(t)$ and $\hat{s}_2(t)$ can be found by using inverse STFT of their corresponding STFTs $\hat {S}_1(t,f)$ and $\hat {S}_2(t,f)$.   
\section{Experiments and Discussion}
We applied the proposed algorithm to separate a speech signal from a background piano music signal. Our main goal was to get a clean speech signal from a mixture of speech and piano signals. We simulated our algorithm on a collection of speech and piano data at 16kHz sampling rate. For speech data, we used the training and testing male speech data from the TIMIT database. For music data, we downloaded piano music data from the piano society web site \citep{site:pianosociety}. We used 12 pieces with approximate 50 minutes total duration from different composers but from a single artist for training and left out one piece for testing. The PSD for the speech and music data were calculated by using the STFT: A Hamming window with 480 points length and $60\%$ overlap was used and the FFT was taken at 512 points, the first 257 FFT points only were used since the conjugate of the remaining 255 points are involved in the first points. We trained 128 basis vectors for each source, which makes the size of $\bB_{\scriptsize {\mbox{speech}}}$ and $\bB_{\scriptsize {\mbox{music}}}$  matrices to be $257 \times 128$, hence, the vector dimension $d=128$ in equation (\ref{GMM}) for both sources. 
The mixed data was formed by adding random portions of the test music file to 20 speech files from the test data of the TIMIT database at different speech-to-music ratio (SMR) values in dB. The audio power levels of each file were found using the ``audio voltmeter'' program from the G.191 ITU-T STL software suite \citep{site:itustl}. For each SMR value, we obtained 20 mixed utterances this way. We used the first 10 utterances as a validation set to choose reasonable values for the regularization parameters $\alpha_{\scriptsize {\mbox{speech}}}$ and $\alpha_{\scriptsize {\mbox{music}}}$ and the number of Gaussian mixture components $K$. The other 10 mixed utterances were used for testing. 

Performance measurement of the separation algorithm was done using the signal to noise ratio (SNR). The average SNR over the 10 test utterances for each SMR case are reported. We also used signal to interference ratio (SIR), which is defined as the ratio of the target energy to the interference error due to the music signal only \citep{vincent:06:pmi}. 

Table \ref{table:alphas_is} shows SNR and SIR of the separated speech signal using NMF with different values of the number of Gaussian mixture components $K$ and fixed regularization parameters $\alpha_{\scriptsize {\mbox{speech}}} = \alpha_{\scriptsize {\mbox{music}}}=1$. The first column of the Table, shows the separation results of using just NMF without any prior.
 
\begin{table}[ht]
\caption{SNR and SIR in dB for the estimated speech signal with regularization parameters $\alpha_{\scriptsize {\mbox{speech}}}=\alpha_{\scriptsize {\mbox{music}}}=1$ and different number of Gaussian mixture components $K$.}
\centering
\scalebox{0.9}
{
\begin{tabular}{||c||c|c||c|c||c|c||c|c||c|c||c|c||}
\hline\hline
SMR & \multicolumn{2}{|c||}{No prior} & \multicolumn{2}{|c||}{$K=1$} & \multicolumn{2}{|c||}{$K=4$}& \multicolumn{2}{|c||}{$K=8$}& \multicolumn{2}{|c||}{$K=16$}& \multicolumn{2}{|c||}{$K=32$} \\
dB  & SNR&SIR & SNR&SIR& SNR&SIR& SNR&SIR& SNR&SIR& SNR&SIR \\        
\hline
-5   & 2.88  & 4.86 & 3.31 & 5.71 & 3.61 & 6.58 &4.24 & 8.07& \bf{4.76} &\bf{10.07}& 4.27 & 8.39 \\
\hline
0    & 5.50  & 8.70 & 5.74 & 9.31 & 5.90 & 9.99 &6.32 & 11.61& 6.45 &\bf{13.02}& \bf{6.54} & 12.42 \\
\hline
5    & 8.37  & 12.20 & 8.46 & 12.40 & 8.55 & 12.98 &\bf{8.74} & 14.13& 8.73 &\bf{15.62}& 8.69 & 14.51 \\
\hline
\hline
\end{tabular}
}
\label{table:alphas_is} 
\end{table} 
As we can see from the Table, the proposed regularized NMF algorithm improves the separation performance for challenging SMR cases compared with using just NMF without priors. Increasing the number of Gaussian mixture components $K$ improves the separation performance until $K=16$. From the shown results, $K=16$ seems to be a good choice for the given data sets. The best choice for $K$ usually depends on the nature and the size of the training data. For example, for speech signal in general there are variety of phonetic differences, gender, speaking styles, accents, which raises the necessity for using many Gaussian components.

\subsection*{Comparison with other priors}

In this section we compared our proposed method of using MMSE estimates under GMM prior on the solution of NMF with two other prior methods. The first prior is the sparsity prior and the second prior is enforced by maximizing the loglikelihood under GMM prior distribution. 

In the sparsity prior, the NMF solution of the gains matrix was enforced to be sparse \citep{virtanen:09:mogpfnnmfbss,Schmidt:06:scssusnnmf}. The sparse NMF is defined as
\begin{equation}
\label{sparse}
C\left(G\right)=D_{IS}\left(\bY\left|\right|\bB\bG\right)+\lambda\sum_{m,n}\bG_{m,n},
\end{equation}
where $\lambda$ is the regularization parameter. The gain update rule of $\bG$ can be found as follows: 
\begin{equation}
\label{weights_sparse}
\bG\leftarrow \bG\otimes\frac{\bB^T\frac{\bY}{\left(\bB\bG\right)^2}}{\bB^T\frac{\Bone}{\bB\bG}+\lambda}.
\end{equation}  
Enforcing sparsity on the NMF solution of the gains matrix is equivalent to model the prior of the gains matrix using exponential distribution with parameter $\lambda$ \citep{virtanen:09:mogpfnnmfbss}. The update rule in equation (\ref{weights_sparse}) is found based on maximizing the likelihood of the gains matrix under the exponential prior distribution.   

The second method of enforcing prior on the NMF solution is by using GMM gain prior \citep{emad:12:gmgprnmfscss,grais:12:rnmfgmpsscss}. The NMF solution for the gains matrix is enforced to increase its log-likelihood with the trained GMM prior as follows:
\begin{equation}
\label{div_sep2}
C=D_{IS}\left(\bY\left|\right|\bB\bG\right)- R_2(\bG),
\end{equation}
where $R_2(\bG)$ is the weighted sum of the log-likelihoods of the log-normalized columns of the gains matrix $\bG$. $R_2(\bG)$ can be written as follows:
\begin{equation}
	R_2(\bG)=\sum_{i=1}^2\eta_i\Gamma(\bG_i), 
\end{equation}
where $\Gamma(\bG_i)$ is the log-likelihood for the submatrix $\bG_i$ for source $i$.
   
In sparsity and GMM based log-likelihood prior methods, to match between the used update rule for the gains matrix during training and separation, the priors were enforced during both training and separation stages. In sparse NMF we used sparsity constraints during training and separation stages. In regularized NMF with GMM based log-likelihood prior we trained the NMF bases and the prior GMM parameters jointly as shown in \citet{grais:12:rnmfgmpsscss}.   

In the sparse NMF case, we got best results when $\lambda=0.0001$ for both sources in the training and separation stages. In the case of enforcing the gains matrix to increase the log-likelihood under GMM prior \citep{grais:12:rnmfgmpsscss} we got the best results when $\eta=1$ in the training and $\eta=0.1$ in the separation stage. The number of Gaussian components was $K=4$ for both sources. It is important to note that, in the case of using MMSE under GMM prior there is no need to enforce prior during training since the uncertainty measurements during training are assumed to be zeros since the training data are clean signals. When the uncertainty is zero, then the regularized NMF in case of MMSE under GMM prior is the same as the NMF cost function, then the update rule for the gains matrix in the training stage is the same as the update rule in the case of using just NMF.           

\begin{figure}[!htb]
  \centering
  \centerline{\includegraphics[width=0.9\textwidth]{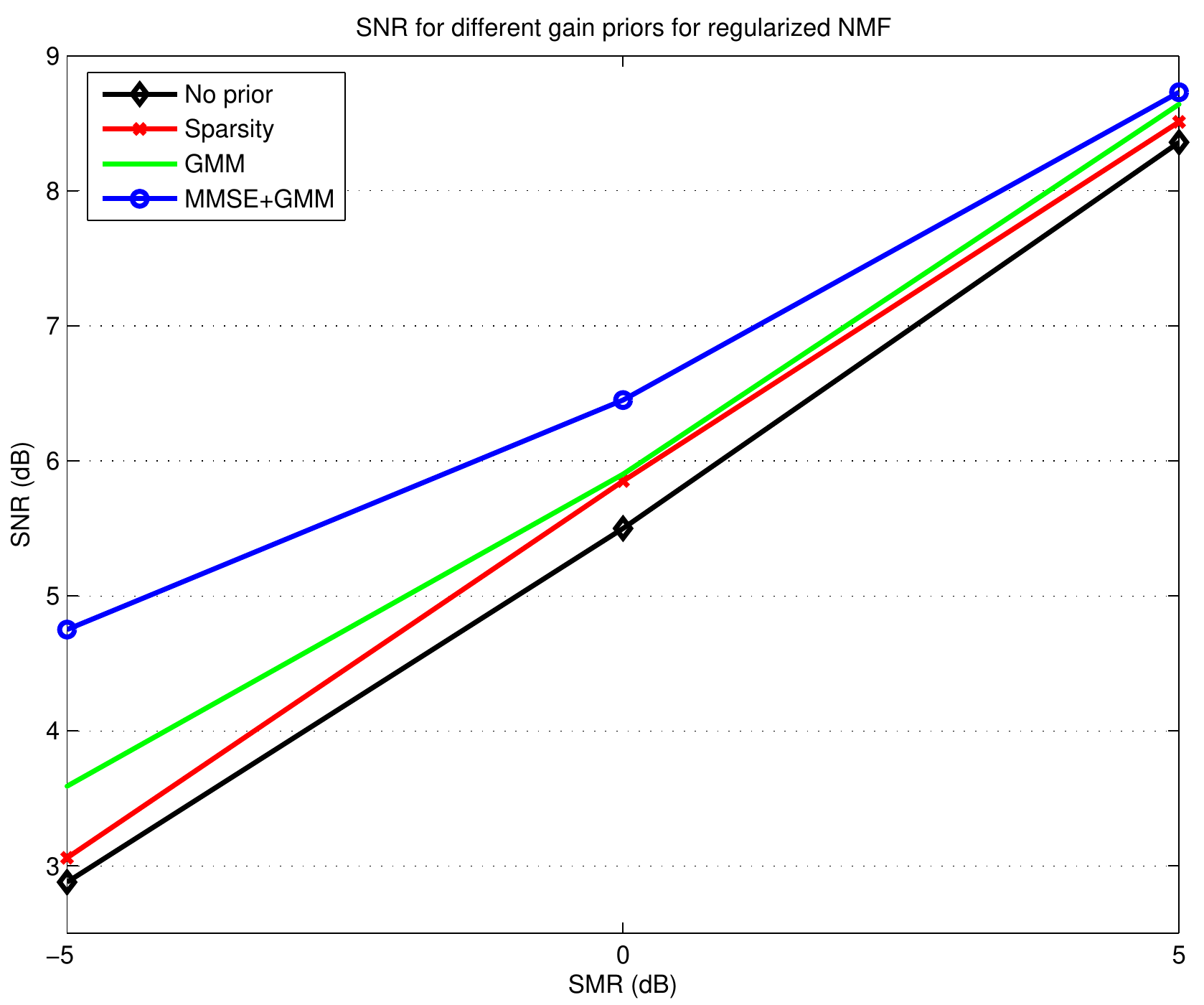}}
%
\caption{The effect of using different prior models on the gains matrix. The black line for using no prior case, the red line for using the exponential distribution prior, the green line is for maximizing the gains matrix likelihood with the GMM prior, and the blue line is for using MMSE under GMM as a prior}
\label{effect_of_k}
\end{figure}

\begin{figure}[!htb]
  \centering
   \centerline{\includegraphics[width=0.9\textwidth]{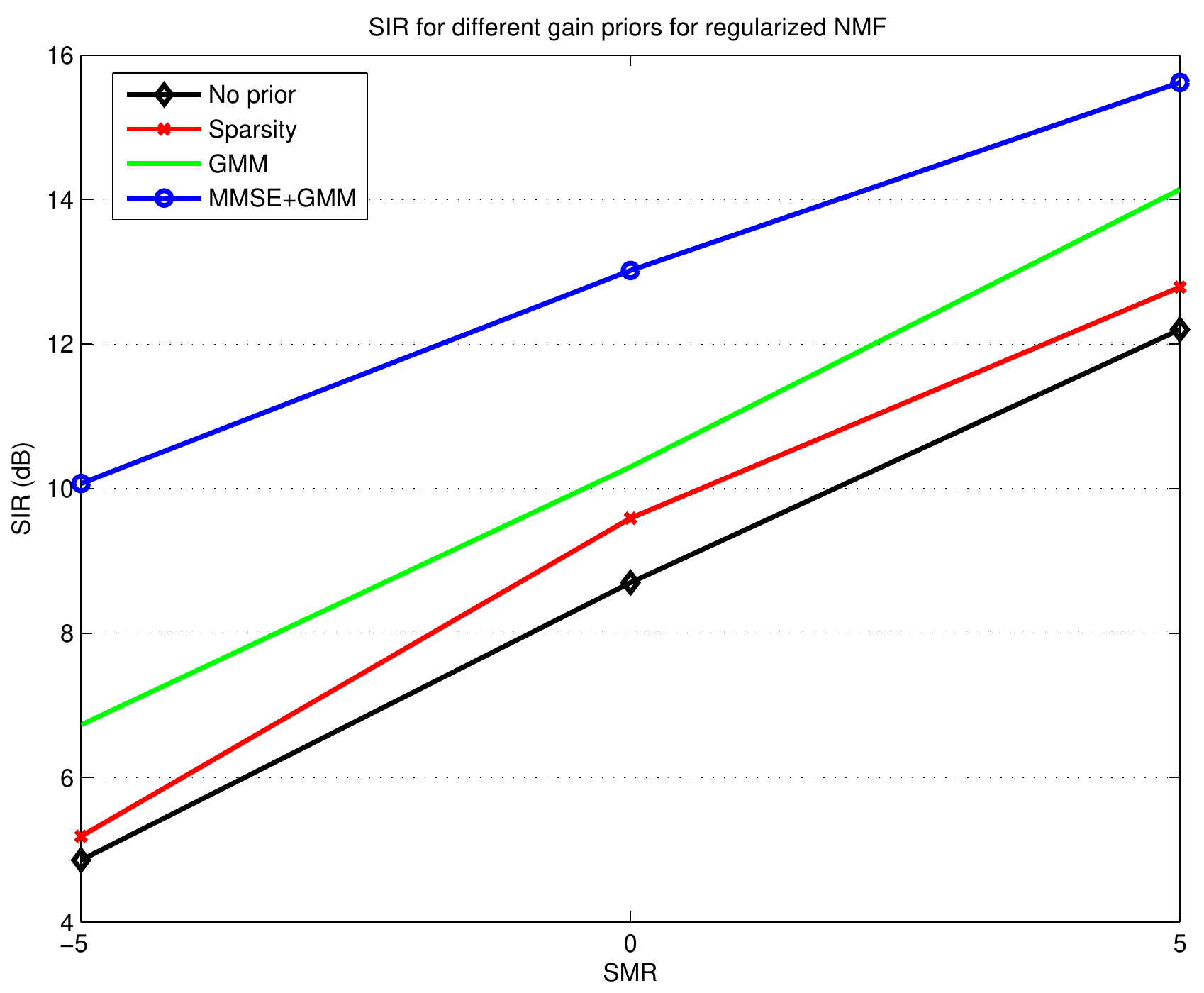}}
\caption{The effect of using different prior models on the gains matrix with the same color map of the previous figure.}
\label{effect_of_k2}
\end{figure}

Figures \ref{effect_of_k} and \ref{effect_of_k2} show the SNR and SIR for the different type of prior models. The black line shows the separation performance in the case of no prior is used. The red line shows the performance for the case of using sparse NMF. The green line shows the performance in the case of enforcing the gains matrix to increase its likelihood with the prior GMM. The blue line shows the separation performance in the case of using MMSE estimate under GMM prior that is proposed in this paper. As we can see, the proposed method of enforcing prior on the gains matrix using MMSE estimate under GMM prior gives the best performance comparing with the other methods. The used MMSE estimates prior in this work gives better results than the GMM likelihood method \citep{grais:12:rnmfgmpsscss} because of the measurements of the uncertainties in the MMSE under GMM case. The uncertainties work as feedback measurements that adjust the needs to the prior based on the amount of distortion in the gains matrix during the separation stage.      

Comparing the relative improvements in dB that we got in this paper with the achieved improvements in other works \citep{Wilson:08:sdunmfwp,kevin:08:srnmfwtdfs,virtanen:09:mogpfnnmfbss,Virtanen:07:msssbnmfwtcasc} we can see that the, improvements in this paper can be considered to be high.        
%

%
%
\section{\uppercase{Conclusion}}
In this work, we introduced a new regularized NMF algorithm. The NMF solution for the gains matrix was guided by the MMSE estimate under a GMM prior where the uncertainty of the observed mixed signal was learned online from the observed data. The proposed algorithm can be extended for better measurements of the distortion in the observed signal by embedding more parameters in equation (\ref{corrupted1}) that can be learned online from the observed signal.

\section{Acknowledgements}
This research is partially supported by Turk Telekom Group Research and Development, project entitled ``Single Channel Source Separation'', grant number 3014-06, support year 2012.






\bibliographystyle{model2-names}
\bibliography{refs}

\subsubsection*{\textbf{APPENDIX A}}
\label{appendxA}
\begin{figure}[!htb]
  \centering
  \centerline{\includegraphics[width=6cm]{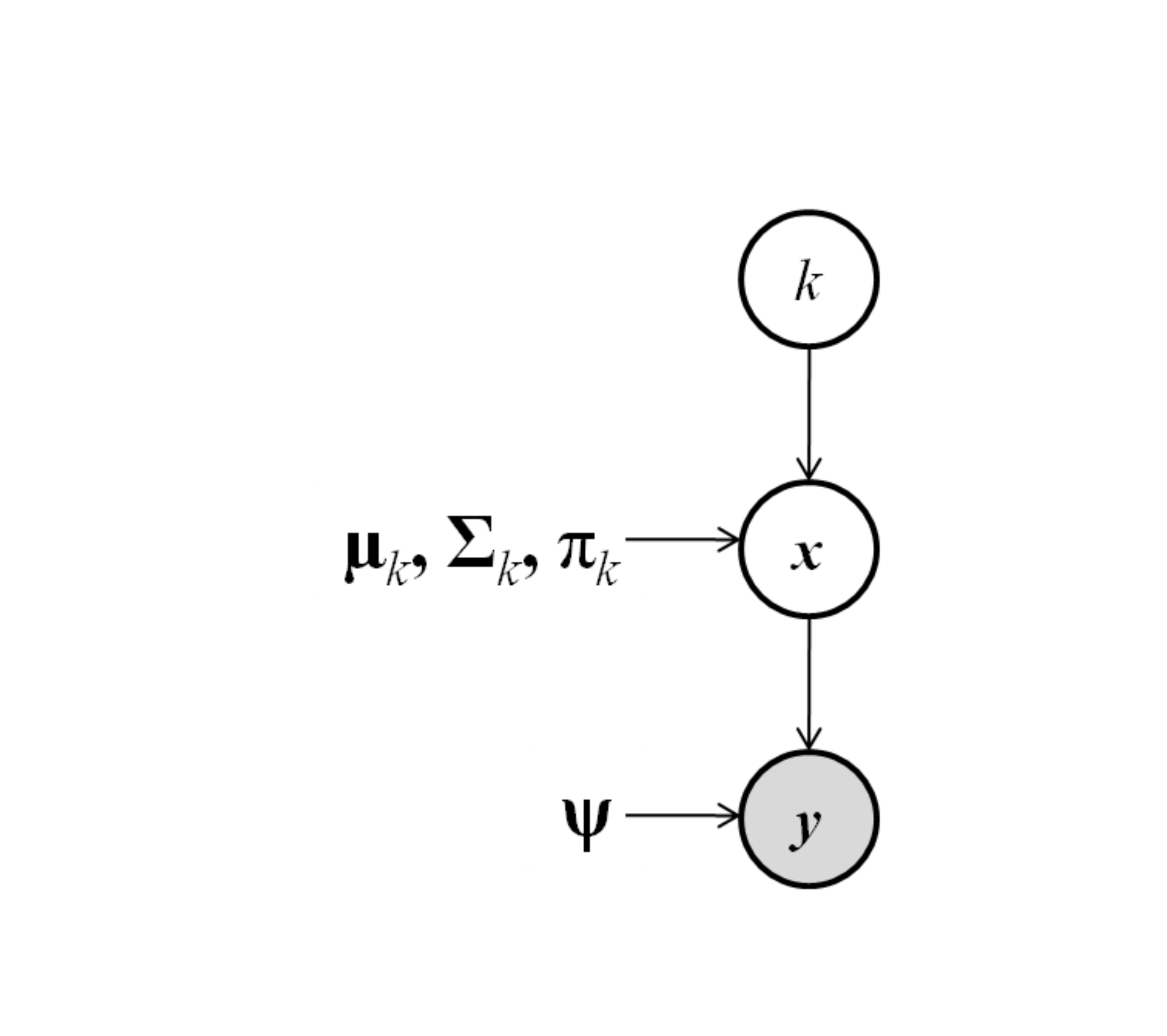}}
%
\caption{The graphical model of the observation model.}
\label{fig:factor_analysis_model}
\end{figure}

In this appendix, we show the MMSE estimate and the parameter $\bPsi$ learning similar to \citet{rosti:01:glgm}, \citet{zoubin:97:emmfa}, and \citet{rosti:04:fahmmfsr}. Assume we have a noisy observation $\By$ as shown in the graphical model in Figure \ref{fig:factor_analysis_model}, which can be formulated as follows:
\begin{equation}
\By=\Bx+\Be,
\end{equation}

where $\Be$ is the noise term, and $\Bx$ is the unknown underlying correct signal which needs to be estimated under a GMM prior distribution: 
\begin{equation}
p\left(\Bx\right) = \sum_{k=1}^K \pi_k \mbox{\rsfsfive{N}} \left(\Bx|\Bmu_k,\bSigma_k\right),
\end{equation}
the error term $\Be$ has a Gaussian distribution with zero mean and diagonal covariance matrix $\bPsi$:
\begin{equation}
p\left(\Be \right) =  \mbox{\rsfsfive{N}} \left(\Be|\Bzero,\bPsi\right).
\end{equation}
The conditional distribution of $\By$ is a Gaussian with mean $\Bx$ and diagonal covariance matrix $\bPsi$:
\begin{equation}
p(\By|\Bx,k)=\mbox{\rsfsfive{N}} \left(\By|\Bx,\bPsi\right).
\end{equation}
The distribution of $\By$ given the Gaussian component $k$ is a Gaussian with mean $\Bmu_k$ and diagonal covariance matrix $\bSigma_k+\bPsi$:
\begin{equation}
p(\By|k)=\mbox{\rsfsfive{N}} \left(\By|\Bmu_k,\bSigma_k+\bPsi\right).
\end{equation}
The marginal probability distribution of $\By$ is a GMM:
\begin{equation}
p(\By)=\sum_{k=1}^K\pi_k \mbox{\rsfsfive{N}} \left(\By|\Bmu_k,\bSigma_k+\bPsi\right),
\end{equation}
where the expectations $E\left(\Bx\right)=E\left(\By\right)$, and $E\left(\Be\right)=\Bzero$.
Note that, this observation model has some mathematical similarities but different concepts with factor analysis models assuming the load matrix is the identity matrix \citep{rosti:01:glgm,zoubin:97:emmfa,rosti:04:fahmmfsr,jordan:01:aitgm}.

The MMSE estimate of $\Bx$ can be found by calculating the conditional expectation of $\Bx$ given the observation $\By$.
Given the Gaussian component $k$, the joint distribution of $\Bx$ and $\By$ is a multivariate Gaussian distribution with conditional expectation and conditional variance as follows \citep{rosti:01:glgm,garcia:94:prpee}:
\begin{equation}
E\left(\Bx|\By,k\right)=\Bmu_k+\bSigma_{k_{\Bx\By}}\bSigma^{-1}_{k_{\By}}\left(\By-\Bmu_k\right),
\end{equation}

\begin{equation}
\mbox{var}\left(\Bx|\By,k\right)=\bSigma_k-\bSigma_{k_{\Bx\By}}\bSigma^{-1}_{k_{\By}}\bSigma_{k_{\Bx\By}}^T,
\end{equation}
we know that 
\begin{equation}
\bSigma_k{_{\By}}=\bSigma_k+\bPsi,
\end{equation}
and
\begin{align}
\bSigma_{k_{\Bx\By}}&=\mbox{cov}\left(\Bx,\By\right)  \nonumber \\
								&=E\left(\Bx\By^T\right)-E\left(\Bx\right)E\left(\By^T\right)  \nonumber \\
								&=E\left[\Bx\left(\Bx^T+\Be^T\right)\right]-E\left(\Bx\right)E\left(\By^T\right)  \nonumber\\
								&=E\left(\Bx\Bx^T\right)+E\left(\Bx\right)E\left(\Be^T\right)-E\left(\Bx\right)E\left(\By^T\right)\nonumber\\
								&=\mbox{var}\left(\Bx\right)+E\left(\Bx\right)E\left(\Bx^T\right)-E\left(\Bx\right)E\left(\By^T\right)\nonumber\\
								&=\mbox{var}\left(\Bx\right)=\bSigma_k.
\end{align}
The conditional expectation given the Gaussian component $k$ of the prior model is

\begin{align}
\label{zk_hat}
E\left(\Bx|\By,k\right)&=\Bmu_k+\bSigma_k\left(\bSigma_k+\bPsi\right)^{-1}\left(\By-\Bmu_k\right) \nonumber\\
											 &=\hat{\Bx}_k.
\end{align}  
We also can find the following conditional expectation given only the observation $\By$ as follows:


\begin{align}
\label{cond_expect} 
E\left(\Bx|\By\right)&=\sum_{k=1}^KE\left(k|\By\right)E\left(\Bx|\By,k\right) \nonumber \\
									 &=\sum_{k=1}^K\gamma_kE\left(\Bx|\By,k\right) \nonumber \\
									 &=\hat {\Bx},
\end{align} 

where
\begin{equation}
\label{respons}
E\left(k|\By\right)=\frac{\pi_kp\left(\By|k\right)}{\sum_{j=1}^K\pi_j p\left(\By|j\right)}=\gamma_k.
\end{equation}

From equations (\ref{zk_hat}, \ref{cond_expect}, \ref{respons}) we can write the final MMSE estimate of $\Bx$ given the model parameters as follows:
\begin{equation}
\label{zk_hat_final3}
\hat {\Bx}=\sum_{k=1}^K\gamma_k\left[\Bmu_k+\bSigma_k\left(\bSigma_k+\bPsi\right)^{-1}\left(\By-\Bmu_k\right)\right].
\end{equation}

We need also to find the following sufficient statistics to be used in estimating the model parameters: 

\begin{equation}
\mbox{var}\left(\Bx|\By,k\right)=\bSigma_k-\bSigma_k\left(\bSigma_k+\Psi\right)^{-1}\bSigma_k^T,
\end{equation}

\begin{align}
\label{corr_z}
E\left(\Bx\Bx^T|\By,k\right)&=\mbox{var}\left(\Bx|\By,k\right)+E\left(\Bx|\By,k\right)E\left(\Bx|\By,k\right)^T  \nonumber \\
														&=\bSigma_k-\bSigma_k\left(\bSigma_k+\bPsi\right)^{-1}\bSigma_k^T+\hat{\Bx}_k\hat{\Bx}_k^T \nonumber \\
														&=\hat{\bR}_k,
\end{align}    
and
\begin{align} 
\label{cond_var}
E\left(\Bx\Bx^T|\By\right)&=\sum_{k=1}^KE\left(k|\By\right)E\left(\Bx\Bx^T|\By,k\right) \nonumber \\
									 &=\sum_{k=1}^K\gamma_kE\left(\Bx\Bx^T|\By,k\right) \nonumber \\
									 &=\sum_{k=1}^K\gamma_k\hat{\bR}_k \nonumber \\
									 &=\hat{\bR}.
\end{align} 

\subsubsection*{Parameters learning using the EM algorithm}
In the training stage, we assume we have clean data with $\Be=\Bzero$. The prior GMM parameters $\pi, \Bmu, \bSigma$ are learned as regular GMM models. The only parameter that need to be estimated is $\bPsi$, which is learned from the deformed signal ``$\Bq_n$'' in the paper. The parameter $\bPsi$ is learned iteratively using maximum likelihood estimation. Given the data points $\Bq=\Bq_1, \Bq_2,.., \Bq_n, ...., \Bq_N$, and the GMM parameters, we need to find an estimate for $\bPsi$. We follow the same procedures as in \citet{rosti:01:glgm}, \citet{zoubin:97:emmfa}, and \citet{rosti:04:fahmmfsr}. 

Lets rewrite the sufficient statistics in equations (\ref{respons}, \ref{zk_hat}, \ref{zk_hat_final3}, \ref{corr_z}, \ref{cond_var}) after replacing $\Bx$ with $\Bz$ (to avoid confusion between calculating MMSE and training the model parameters) as follows:
\begin{equation}
\label{gamma_training_final}
\gamma_{kn}=\frac{\pi_k \mbox{\rsfsfive{N}} \left(\Bq_n|\Bmu_k,\bSigma_k+\bPsi \right)}{\sum_{j=1}^K \pi_j \mbox{\rsfsfive{N}} \left(\Bq_n|\Bmu_j,\bSigma_j+\bPsi \right)},
\end{equation} 

\begin{equation}
\label{cond_expect_train}
\hat{\Bz}_{kn}=E\left(\Bz|\Bq_n,k\right)=\Bmu_k+\bSigma_k\left(\bSigma_k+\bPsi\right)^{-1}\left(\Bq_n-\Bmu_k\right),
\end{equation} 

\begin{equation}
\label{cond_expect_train_final}
\hat{\Bz}_{n}=E\left(\Bz|\Bq_n\right)=\sum_{k=1}^K \gamma_{kn}\hat{\Bz}_{kn},
\end{equation}

\begin{equation}
\label{cond_corr_train}
\hat{\bR}_{kn}=E\left(\Bz\Bz^T|\Bq_n,k\right)=\bSigma_k-\bSigma_k\left(\bSigma_k+\bPsi\right)^{-1}\bSigma_k^T+\hat{\Bz}_{kn}\hat{\Bz}_{kn}^T,
\end{equation} 
and
\begin{equation}
\label{cond_corr_train_final}
\hat{\bR}_n=E\left(\Bz\Bz^T|\Bq_n\right)=\sum_{k=1}^K \gamma_{kn}\hat{\bR}_{kn}.
\end{equation}
The complete log-likelihood can be written in a product form as follows:
\begin{align}
\label{loglikelihood_train}
l\left(\Bq,\Bz,k | \Bmu,\bSigma,\pi,\bPsi\right)&=\log\prod^N_{n=1}\prod^K_{k=1}p(k)p(\Bz|k)p(\Bq_n|\Bz,k), \nonumber\\	&=\log\prod^N_{n=1}\prod^K_{k=1}\left[\pi_k \mbox{\rsfsfive{N}} \left(\Bz|\Bmu_k,\bSigma_k\right) \mbox{\rsfsfive{N}} \left(\Bq_n|\Bz,\bPsi\right)\right]^k,
\end{align}

\begin{equation}
l\left(\Bq,\Bz,k | \Bmu,\bSigma,\pi,\bPsi\right)=\sum^N_{n=1}\sum^K_{k=1}k\log \pi_k+\sum^N_{n=1}\sum^K_{k=1}k\log  \mbox{\rsfsfive{N}} \left(\Bz|\Bmu_k,\bSigma_k\right)+\sum^N_{n=1}\sum^K_{k=1}k \log \mbox{\rsfsfive{N}} \left(\Bq_n|\Bz,\bPsi\right).
\end{equation}

The conditional expectation of the complete log likelihood, which is conditioning on the observed data $\Bq_n$ can be written as:

\begin{align}
\label{auxiolary}
Q  =&\sum^N_{n=1}\sum^K_{k=1}E_{\Bq_n}\left(k|\Bq_n\right)\log \pi_k+\sum^N_{n=1}\sum^K_{k=1}E_{\Bq_n}\left(k|\Bq_n\right)E_{\Bq_n}\left(\log \mbox{\rsfsfive{N}} \left(\Bz|\Bmu_k,\bSigma_k\right)|\Bq_n\right) \nonumber \\
&+\sum^N_{n=1}\sum^K_{k=1}E_{\Bq_n}\left(k|\Bq_n\right) E_{\Bq_n}\left(\log \mbox{\rsfsfive{N}} \left(\Bq_n|\Bz,\bPsi\right)|\Bq_n\right),
\end{align}
given that
\begin{equation}
E_{\Bq_n}\left(k|\Bq_n\right)=\frac{\pi_k \mbox{\rsfsfive{N}} \left(\Bq_n|\Bmu_k,\bSigma_k+\bPsi \right)}{\sum_{j=1}^K \pi_j \mbox{\rsfsfive{N}} \left(\Bq_n|\Bmu_j,\bSigma_j+\bPsi \right)}=\gamma_{kn}. 
\end{equation}
We can write the complete log-likelihood as follows:
\begin{align}
\label{auxiolary2}
Q=&\sum^N_{n=1}\sum^K_{k=1}\gamma_{kn}\log \pi_k+\sum^N_{n=1}\sum^K_{k=1}\gamma_{kn}E_{\Bq_n}\left(\log \mbox{\rsfsfive{N}} \left(\Bz|\Bmu_k,\bSigma_k\right)|\Bq_n\right) \nonumber \\
&+\sum^N_{n=1}\sum^K_{k=1}\gamma_{kn}E_{\Bq_n}\left(\log \mbox{\rsfsfive{N}} \left(\Bq_n|\Bz,\bPsi\right)|\Bq_n\right).
\end{align}

For the parameter $\bPsi$, we need to maximize the third part of equation (\ref{auxiolary2}) with respect to $\bPsi$: 
\begin{align}
\label{Q_observation}
Q_{\Bq_n}=&\sum^N_{n=1}\sum^K_{k=1}\gamma_{kn}E_{\Bq_n}\left(\log \mbox{\rsfsfive{N}} \left(\Bq_n|\Bz,\bPsi\right)|\Bq_n\right) \nonumber \\
=&\sum^N_{n=1}\sum^K_{k=1}\gamma_{kn} E_{\Bq_n}\left(\log\frac{1}{\left(2\pi\right)^{\frac{d}{2}}\left|\bPsi\right|^{\frac{1}{2}}}\exp\left\{\frac{-1}{2}\left(\Bq_n-\Bz\right)^T\bPsi^{-1}\left(\Bq_n-\Bz\right)\right\}|\Bq_n,k \right), \nonumber \\
=&\sum^N_{n=1}\sum^K_{k=1}\gamma_{kn} E_{\Bq_n}\left(\frac{-d}{2}\log\left(2\pi\right)-\frac{1}{2}\log \left|\bPsi\right|-\frac{1}{2}\left(\Bq_n-\Bz\right)^T\bPsi^{-1}\left(\Bq_n-\Bz\right)|\Bq_n,k \right),
\end{align}
the derivative of $Q_{\Bq_n}$ with respect to $\bPsi^{-1}$ is set to zero:
\begin{equation}
\frac{\partial Q_{\Bq_n}}{\partial \bPsi^{-1}}=\sum^N_{n=1}\sum^K_{k=1}\gamma_{kn}E_{\Bq_n}\left(\frac{1}{2}\bPsi-\frac{1}{2}\left(\Bq_n-\Bz\right)\left(\Bq_n-\Bz\right)^T|\Bq_n,k \right)=\Bzero,
\end{equation}

\begin{align}
\bPsi\sum^N_{n=1}\sum^K_{k=1}\gamma_{kn}=&\sum^N_{n=1}\sum^K_{k=1}\gamma_{kn}\Bq_n\Bq_n^T-\sum^N_{n=1}\Bq_n\sum^K_{k=1}\gamma_{kn}E_{\Bq_n}\left(\Bz|\Bq_n,k\right)^T \nonumber \\
&-\left(\sum^N_{n=1}\Bq_n\sum^K_{k=1}\gamma_{kn}E_{\Bq_n}\left(\Bz|\Bq_n,k\right)^T\right)^T+\sum^N_{n=1}\sum^K_{k=1}\gamma_{kn}E_{\Bq_n}\left(\Bz\Bz^T|\Bq_n,k\right),
\end{align}

we know that
	\[\sum^N_{n=1}\sum^K_{k=1}\gamma_{kn}=N \ \ \ \mbox{and} \ \ \  \sum^K_{k=1}\gamma_{kn}=1,\]
	then
\[\sum^N_{n=1}\Bq_n\Bq_n^T\sum^K_{k=1}\gamma_{kn}=\sum^N_{n=1}\Bq_n\Bq_n^T,\] and \[\bPsi\sum^N_{n=1}\sum^K_{k=1}\gamma_{kn}=N\bPsi.\] 
We can use the values of $\sum^K_{k=1}\gamma_{kn}E_{\Bq_n}\left(\Bz|\Bq_n,k\right)$ and $\sum^K_{k=1}\gamma_{kn}E_{\Bq_n}\left(\Bz\Bz^T|\Bq_n,k\right)$ from equations (\ref{cond_expect_train_final}, \ref{cond_corr_train_final}) to find the estimate of $\bPsi$	as follows:

\begin{equation}
\label{epsi_estmate_train}
\hat{\bPsi}=\mbox{diag}\left\{\frac{1}{N}\sum^N_{n=1}\left(\Bq_n\Bq_n^T-\Bq_n\hat{z}_n^T-\hat{z}_n\Bq_n^T+\hat{R}_n\right)   \right\},
\end{equation}
where the ``diag'' operator sets all the off-diagonal elements of a matrix to zero. 

\subsubsection*{\textbf{APPENDIX B}}
\label{appendxB}
In this appendix, we show the gradients of the penalty term in the regularized NMF cost function in section 2.1. To calculate the update rule for the gains matrix $\bG$, the gradients $\nabla_G^+L(\bG)$ and $\nabla_G^-L(\bG)$ are needed to be calculated. Lets recall the regularized NMF cost function
\begin{equation}
\label{div_sep}
C\left(G\right)=D_{IS}\left(\bV\left|\right|\bB\bG\right)+ \alpha L(\bG),
\end{equation}   
where
\begin{equation}
\label{first_constr}
L(\bG)=\sum_{n}^N \left\|\frac{\Bg_{n}}{\left\|\Bg_{n}\right\|_2}-\exp\left(f\left(\Bg_n\right) \right)\right\|_2^2, 
\end{equation}

\begin{equation}
\label{gool}
f\left(\Bg_n\right)=\sum_{k=1}^K\gamma_{k_n}\left[\Bmu_k+\bSigma_k\left(\bSigma_k+\bPsi\right)^{-1}\left(\log\frac{\Bg_{n}}{\left\|\Bg_{n}\right\|_2}-\Bmu_k\right)\right],
\end{equation}
and
\begin{equation}
\gamma_{k_n}=\left[\frac{\pi_k \mbox{\rsfsfive{N}} \left(\log\frac{\Bg_{n}}{\left\|\Bg_{n}\right\|_2}|\Bmu_k,\bSigma_k+\bPsi \right)}{\sum_{j=1}^K \pi_j \mbox{\rsfsfive{N}} \left(\log\frac{\Bg_{n}}{\left\|\Bg_{n}\right\|_2}|\Bmu_j,\bSigma_j+\bPsi \right)}\right].
\end{equation}
Since the training data for the GMM models are the logarithm of the normalized vectors, then the mean vectors of the GMM are always not positive, also the values of $\log\frac{\Bg_{n}}{\left\|\Bg_{n}\right\|_2}$ are also not positive, and $\Bg_n$ is always nonnegative. 

Let $\Bg_n=\Bx$, and its component $a$ is $\Bg_{n_a}=x_a$, and $f(\Bg_n)=f(\Bx)$. 
We can write the constraint in equation (\ref{first_constr}) as:
\begin{equation}
\label{first_constr2}
L(\Bx)=\left\|\frac{\Bx}{\left\|\Bx\right\|_2}-\exp\left(f(\Bx)\right)\right\|_2^2. 
\end{equation}
The $a$ component of the gradient of $L(\Bx)$ is
\begin{align} 
\label{J_driv12}
\frac{\partial L(\Bx)}{\partial x_a}&=2\left(  \frac{x_a}{\left\|\Bx\right\|_2}-\exp\left(f(x_a)\right) \right) \left( \frac{1}{\left\|\Bx\right\|_2}-\frac{x_a^2}{\left\|\Bx\right\|_2^3}-\nabla f(x_a)\exp\left(f(x_a)\right)\right) \nonumber \\
&=\nabla L(x_a), 
\end{align} 
which can be written as a difference of two positive terms
\begin{equation}
\label{J_driv223}
\nabla L(x_a)=\nabla^+ L(x_a)-\nabla^- L(x_a). 
\end{equation} 
The component $a$ of the gradient of $f\left(\Bx\right)$ can be written as a difference of two positive terms:
\begin{equation}
\label{devf_x}
\frac{\partial f\left(\Bx\right)}{\partial x_a}=\nabla^+ f\left(x_a\right)-\nabla^- f\left(x_a\right). 
\end{equation}
The component $a$ of the gradient of $L\left(\Bx\right)$ in equation (\ref{J_driv223}) can be written as:
\begin{equation}
\label{J_driv_pos2}
\footnotesize{
\nabla^+ L(x_a)=2\left\{\frac{x_a}{\left\|\Bx\right\|_2}\left(\frac{1}{\left\|\Bx\right\|_2}+\exp\left(f(x_a)\right)\nabla^- f(x_a) \right)+\exp\left(f(x_a)\right)\left(\frac{x_a^2}{\left\|\Bx\right\|_2^3}+\exp\left(f(x_a)\right)\nabla^+ f(x_a)\right)\right\},
}
\end{equation}
and
\begin{equation}
\label{J_driv_neg2}
\footnotesize{
\nabla^- L(x_a)=2\left\{\frac{x_a}{\left\|\Bx\right\|_2}\left(\frac{x_a^2}{\left\|\Bx\right\|_2^3}+\exp\left(f(x_a)\right)\nabla^+ f(x_a)\right)+\exp\left(f(x_a)\right)\left(\frac{1}{\left\|\Bx\right\|_2}+\exp\left(f(x_a)\right)\nabla^- f(x_a) \right)\right\}.
}
\end{equation}

We need to find the values of $\nabla^+f(x_a)$ and $\nabla^-f(x_a)$. Note that, the term $\bSigma_k\left(\bSigma_k+\bPsi\right)^{-1}$ forms a diagonal matrix. 

Let 
\begin{equation}
\label{total_H}
H(x_a)=\Bmu_{k_a}+\bSigma_{k_{aa}}\left(\bSigma_{k_{aa}}+\bPsi_{aa}\right)^{-1}\left(\log\frac{x_a}{\left\|\Bx\right\|_2}-\Bmu_{k_a}\right),
\end{equation}
then $f(\Bx)$ in equation (\ref{gool}) can be written as:
\begin{equation}
\label{total_fH}
f(\Bx)=\sum_{k=1}^K\gamma_k(\Bx)H(\Bx).
\end{equation}
The gradient of $f(\Bx)$ in equation (\ref{total_fH}) can be written as:
\begin{equation}
\label{fx_derv}
\nabla f(x_a)=\sum_{k=1}^K\left[\gamma_k(\Bx)\nabla H(x_a)+H(x_a)\nabla \gamma_k(x_a)\right],
\end{equation}
where
\begin{equation}
\label{gamma_x}
\gamma_k(\Bx)=\left[\frac{\pi_k \mbox{\rsfsfive{N}} \left(\log\frac{\Bx}{\left\|\Bx\right\|_2}|\Bmu_k,\bSigma_k+\bPsi \right)}{\sum_{j=1}^K \pi_j \mbox{\rsfsfive{N}} \left(\log\frac{\Bx}{\left\|\Bx\right\|_2}|\Bmu_j,\bSigma_j+\bPsi\right)}\right]=\frac{M_k(\Bx)}{N_k(\Bx)}.
\end{equation}
We can also write the gradient components of $H(x_a)$ and $\gamma_k(\Bx)$ as a difference of two positive terms
\begin{equation}
\label{H_derv}
\nabla H(x_a)=\nabla^+ H(x_a)- \nabla^- H(x_a),
\end{equation}
and
\begin{equation}
\label{gamma_totl_derv}
\nabla \gamma_k(x_a)=\nabla^+ \gamma_k(x_a)- \nabla^- \gamma_k(x_a).
\end{equation}

The gradient of $f(x_a)$ in equations (\ref{devf_x}, \ref{fx_derv}) can be written as:
\begin{equation}
\label{fx_derv_pos}
\nabla^+ f(x_a)=\sum_{k=1}^K\left[\gamma_k(\Bx)\nabla^+H(x_a)+H^+(x_a)\nabla^+ \gamma_k(x_a)+H^-(x_a)\nabla^- \gamma_k(x_a)\right],
\end{equation}

\begin{equation}
\label{fx_derv_neg}
\nabla^- f(x_a)=\sum_{k=1}^K\left[\gamma_k(\Bx)\nabla^-H(x_a)+H^-(x_a)\nabla^+ \gamma_k(x_a)+H^+(x_a)\nabla^- \gamma_k(x_a)\right],
\end{equation}
where

\begin{equation}
\label{H_derv_pos}
\nabla^+ H(x_a)=\bSigma_{k_{aa}}\left(\bSigma_{k_{aa}}+\bPsi_{aa}\right)^{-1}\frac{1}{x_a},
\end{equation}
\begin{equation}
\label{H_derv_neg}
\nabla^- H(x_a)=\bSigma_{k_{aa}}\left(\bSigma_{k_{aa}}+\bPsi_{aa}\right)^{-1}\frac{x_a}{\left\|\Bx\right\|_2^2},
\end{equation}
and $H(x_a)$ can be written as a difference of two positive terms:
\begin{equation}
\label{two_H}
H(x_a)=H^+(x_a)-H^-(x_a),
\end{equation}
where
\begin{equation}
\label{postv_H}
H^+(x_a)=-\bSigma_{k_{aa}}\left(\bSigma_{k_{aa}}+\bPsi_{aa}\right)^{-1}\Bmu_{k_a},
\end{equation}
and
\begin{equation}
\label{negtv_H}
H^-(x_a)=-\left[ \Bmu_{k_a}+\bSigma_{k_{aa}}\left(\bSigma_{k_{aa}}+\bPsi_{aa}\right)^{-1}\log\frac{x_a}{\left\|\Bx\right\|_2} \right].
\end{equation}
We can rewrite $\gamma_k(\Bx)$ in equation (\ref{gamma_x}) as:
\begin{equation}
\label{gamma_x2}
\gamma_k(\Bx)=\frac{M_k(\Bx)}{N_k(\Bx)},
\end{equation}
note that $\gamma_k(\Bx), M_k(\Bx), N_k(\Bx)\geq 0$.

The component $a$ of the gradient of $\gamma_k(\Bx)$ can be written as:
\begin{equation}
\label{gamma_derv}
\nabla \gamma_k(x_a)=\frac{N_k(\Bx)\nabla M_k(x_a)-M_k(\Bx)\nabla N_k(x_a)}{N_k^2(\Bx)}.
\end{equation}
We can write the gradients of $M_k(\Bx)$ and $N_k(\Bx)$ as a difference of two positive terms 
\begin{equation}
\label{M_totl_derv}
\nabla M_k(x_a)=\nabla^+ M_k(x_a)- \nabla^- M_k(x_a),
\end{equation}
and
\begin{equation}
\label{N_totl_derv}
\nabla N_k(x_a)=\sum_{k=1}^K\nabla^+ M_k(x_a)-\sum_{k=1}^K \nabla^- M_k(x_a).
\end{equation}
The gradient of $\gamma_k(x_a)$ in equation (\ref{gamma_totl_derv}) can be written as:
\begin{equation}
\label{gamma_derv_pos}
\nabla^+ \gamma_k(x_a)=\frac{N_k(\Bx)\nabla M^+_k(x_a)+M_k(\Bx)\sum_{k=1}^K \nabla^- M_k(x_a)}{N_k^2(\Bx)},
\end{equation}

\begin{equation}
\label{gamma_derv_neg}
\nabla^- \gamma_k(x_a)=\frac{N_k(\Bx)\nabla M^-_k(x_a)+M_k(\Bx)\sum_{k=1}^K \nabla^+ M_k(x_a)}{N_k^2(\Bx)},
\end{equation}
where
\begin{equation}
\label{M_derv_pos}
\nabla^+ M_k(x_a)=M_k(\Bx)\left(\bSigma_{k_{aa}}+\bPsi_{aa}\right)^{-1}\left[\frac{-1}{x_a}\log\frac{x_a}{\left\|\Bx\right\|_2}-\frac{\Bmu_{k_a} x_a}{\left\|\Bx\right\|^2_2}\right],
\end{equation}
and
\begin{equation}
\label{M_derv_neg}
\nabla^- M_k(x_a)=M_k(\Bx)\left(\bSigma_{k_{aa}}+\bPsi_{aa}\right)^{-1}\left[\frac{-\Bmu_{k_a}}{x_a}-\frac{x_a}{\left\|\Bx\right\|^2_2}  \log\frac{x_a}{\left\|\Bx\right\|_2}\right].
\end{equation}
After finding $\nabla^+ \gamma_k(x_a)$, and $\nabla^- \gamma_k(x_a)$ from equations (\ref{gamma_derv_pos}, \ref{gamma_derv_neg}), and $\nabla^+ H(x_a)$, and $\nabla^- H(x_a)$ from equations (\ref{H_derv_pos}, \ref{H_derv_neg}), we can find the gradients $\nabla^+ f(x_a)$, and $\nabla^- f(x_a)$ in equations (\ref{fx_derv_pos}, \ref{fx_derv_neg}), which complete our solution for $\nabla^+ L(x_a)$, and $\nabla^- L(x_a)$ in equations (\ref{J_driv_pos2}, \ref{J_driv_neg2}).








\end{document}

%% file: def_bold.tex
\providecommand{\myboldg}[1]    {\mbox{\boldmath $#1$}}         
\providecommand{\myboldG}       {\myboldg}                      
\providecommand{\mybolda}       {\myboldg}                      
\providecommand{\myboldA}       {\mybolda}                      
\providecommand{\myboldn}       {\mybolda}                      

\providecommand{\Bzero} {\myboldn{0}}
\providecommand{\Bone}  {\myboldn{1}}

\providecommand{\Be}{\mybolda{e}}

\providecommand{\Bg}{\mybolda{g}}

\providecommand{\Bq}{\mybolda{q}}

\providecommand{\Bx}{\mybolda{x}}
\providecommand{\By}{\mybolda{y}}
\providecommand{\Bz}{\mybolda{z}}

\providecommand{\bB}{\myboldA{B}}

\providecommand{\bE}{\myboldA{E}}

\providecommand{\bG}{\myboldA{G}}
\providecommand{\bH}{\myboldA{H}}

\providecommand{\bR}{\myboldA{R}}
\providecommand{\bS}{\myboldA{S}}

\providecommand{\bV}{\myboldA{V}}

\providecommand{\bY}{\myboldA{Y}}

\providecommand{\Bmu}           {\myboldg{\mu}}

\providecommand{\bPsi}          {\myboldG{\Psi}}

\providecommand{\bSigma}        {\myboldG{\Sigma}}